%% file: main.tex
\documentclass[10pt,twocolumn,letterpaper]{article}

\usepackage{iccv}
\usepackage{times}
\usepackage{epsfig}
\usepackage{graphicx}
\usepackage{amsmath}
\usepackage{amssymb}

\input{packages.tex}

\usepackage[pagebackref, colorlinks, breaklinks=true, bookmarks=false]{hyperref}
\usepackage[capitalize]{cleveref}

\hypersetup{
    pdftitle={The TYC Dataset for Understanding Instance-Level Semantics and Motions of Cells in Microstructures},
    pdfsubject={TYC Dataset},
    pdfauthor={Christoph Reich, Tim Prangemeier, and Heinz Koeppl}
}

\newcommand \colorindicator[1]{%
	{\textcolor{#1}{$\blacksquare\!\!\!\!\blacksquare$}}%
}

\renewcommand*{\paragraph}[1]{\smallskip\noindent\textbf{#1}\hspace{0.3em}}

\usepackage[capitalize]{cleveref}
\crefname{section}{Sec.}{Secs.}
\Crefname{section}{Section}{Sections}
\Crefname{table}{Table}{Tables}
\crefname{table}{Tab.}{Tabs.}

\input{colors.tex}

\DeclareSIUnit\fps{fps}

\def\aspace{\hspace{1.5cm}}

\iccvfinalcopy 

\def\iccvPaperID{--} 

\usepackage{fancyhdr}
\usepackage{setspace}

\fancypagestyle{firststyle}
{

    \fancyhf{}
    \lfoot{{\footnotesize\begin{spacing}{.5}\parbox{\linewidth}{\vspace{2em}
    To appear in Proceedings of the \emph{IEEE/CVF International Conference on Computer Vision Workshops (ICCVW)}, BioImage Computing @ ICCV, 2023.%
    \\\hrule\vspace{\baselineskip}
    \copyright~2023 IEEE. Personal use of this material is permitted. Permission from IEEE must be obtained for all other uses, in any current or future media, including reprinting/republishing this material for advertising or promotional purposes, creating new collective works, for resale or redistribution to servers or lists, or reuse of any copyrighted component of this work in other works. 
    }\end{spacing}}}
}

\begin{document}

\title{The TYC Dataset for Understanding Instance-Level Semantics and Motions of\\ Cells in Microstructures}

\author{Christoph Reich
\aspace Tim Prangemeier
\aspace Heinz Koeppl\\
Department of Electrical Engineering and Information Technology, Centre for Synthetic Biology,\\
Technische Universität Darmstadt\\
{\tt\small \{christoph.reich, tim.prangemeier, heinz.koeppl\}@bcs.tu-darmstadt.de}\\
{\tt\small \url{https://christophreich1996.github.io/tyc_dataset}}
}

{
\twocolumn[{
\renewcommand\twocolumn[1][]{#1}
\maketitle
\begin{center}
    \sffamily
    \vspace{-1.5em}
    \includegraphics[width=0.184\linewidth, trim={0 0cm 0 0cm}, clip, frame]{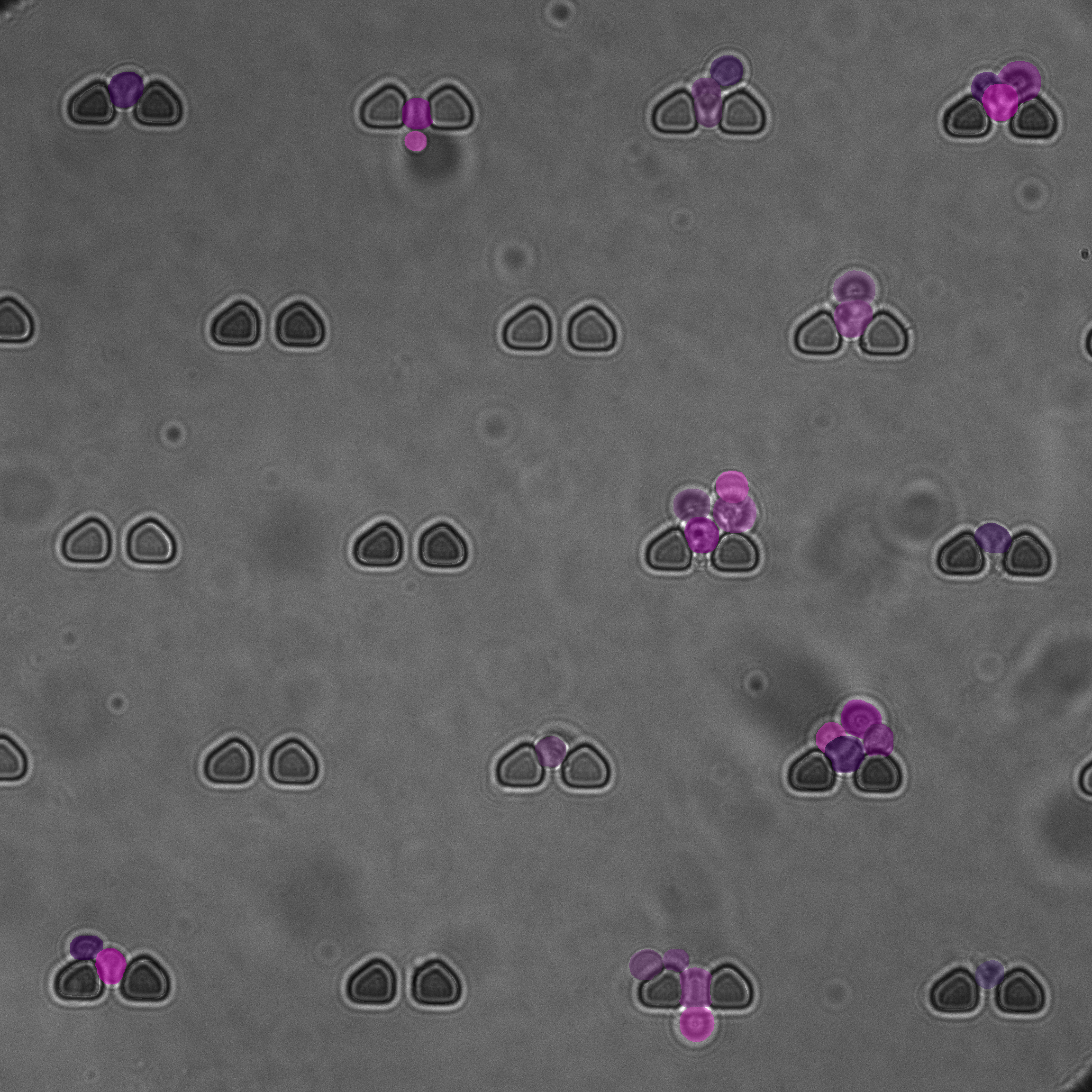}\hfill%
    \includegraphics[width=0.184\linewidth, trim={0 0cm 0 0cm}, clip, frame]{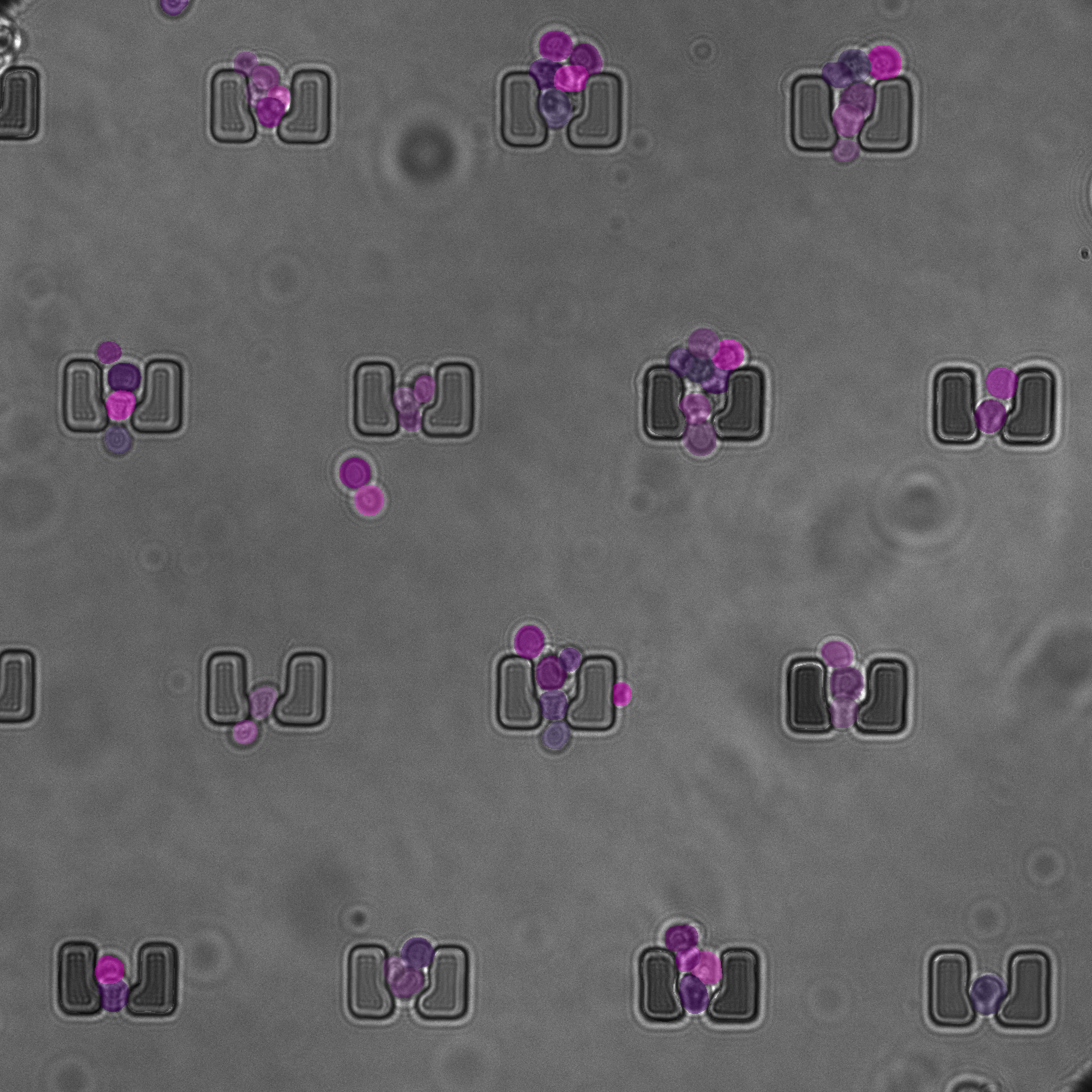}\hfill%
    \includegraphics[width=0.184\linewidth, trim={0 0cm 0 0cm}, clip, frame]{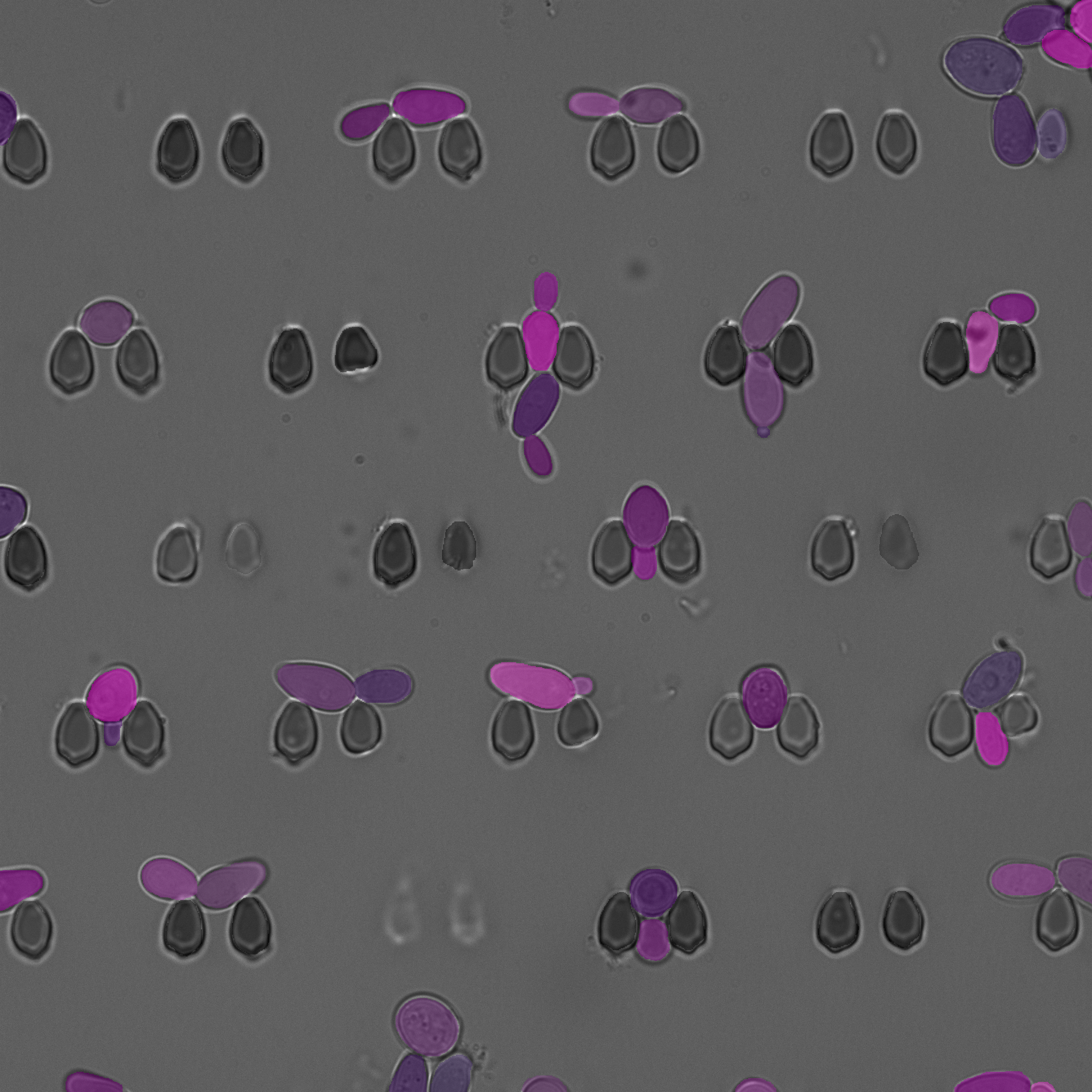}\hfill%
    \includegraphics[width=0.184\linewidth, trim={0 0cm 0 0cm}, clip, frame]{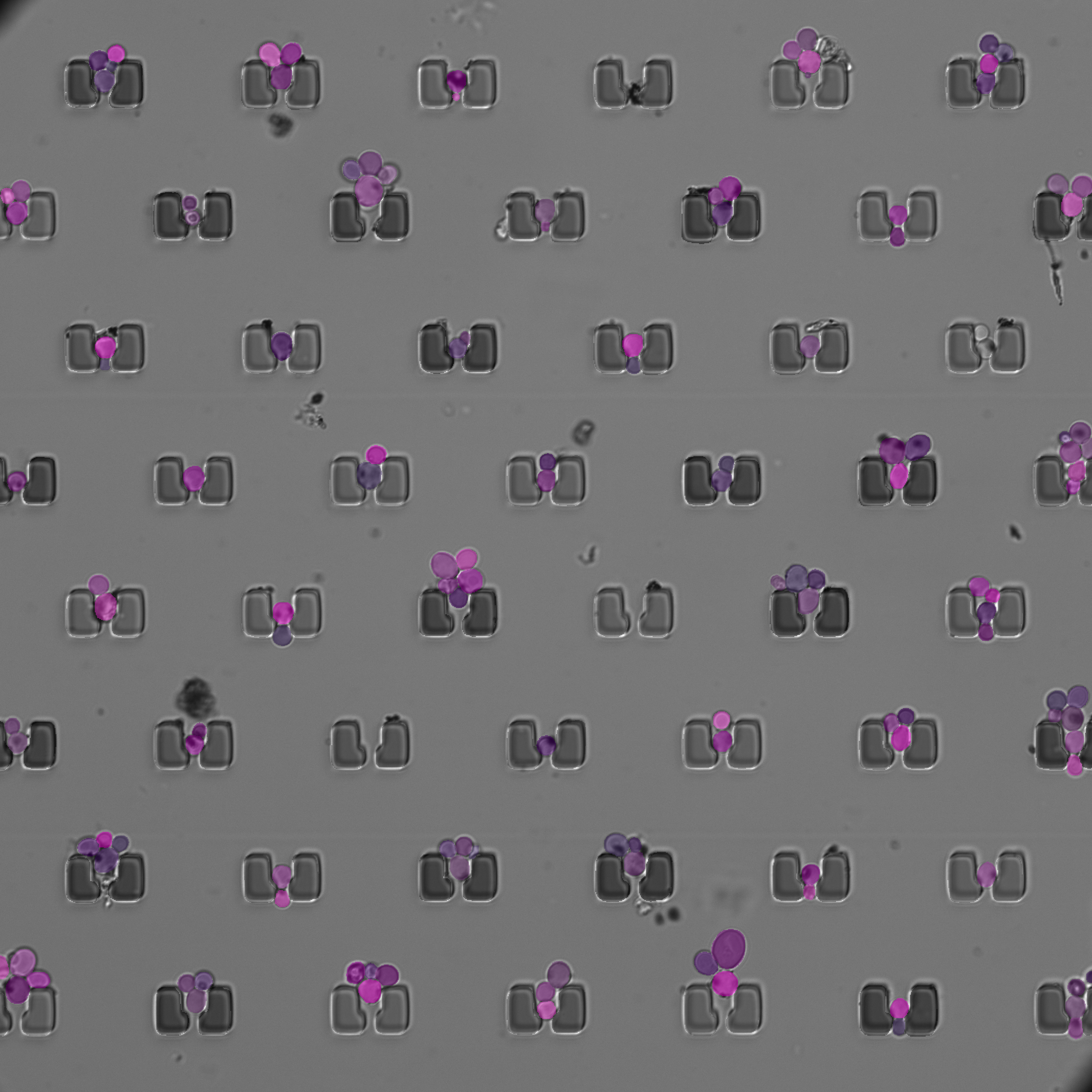}\hfill%
     \includegraphics[width=0.184\linewidth, trim={0 0cm 0 0cm}, clip, frame]{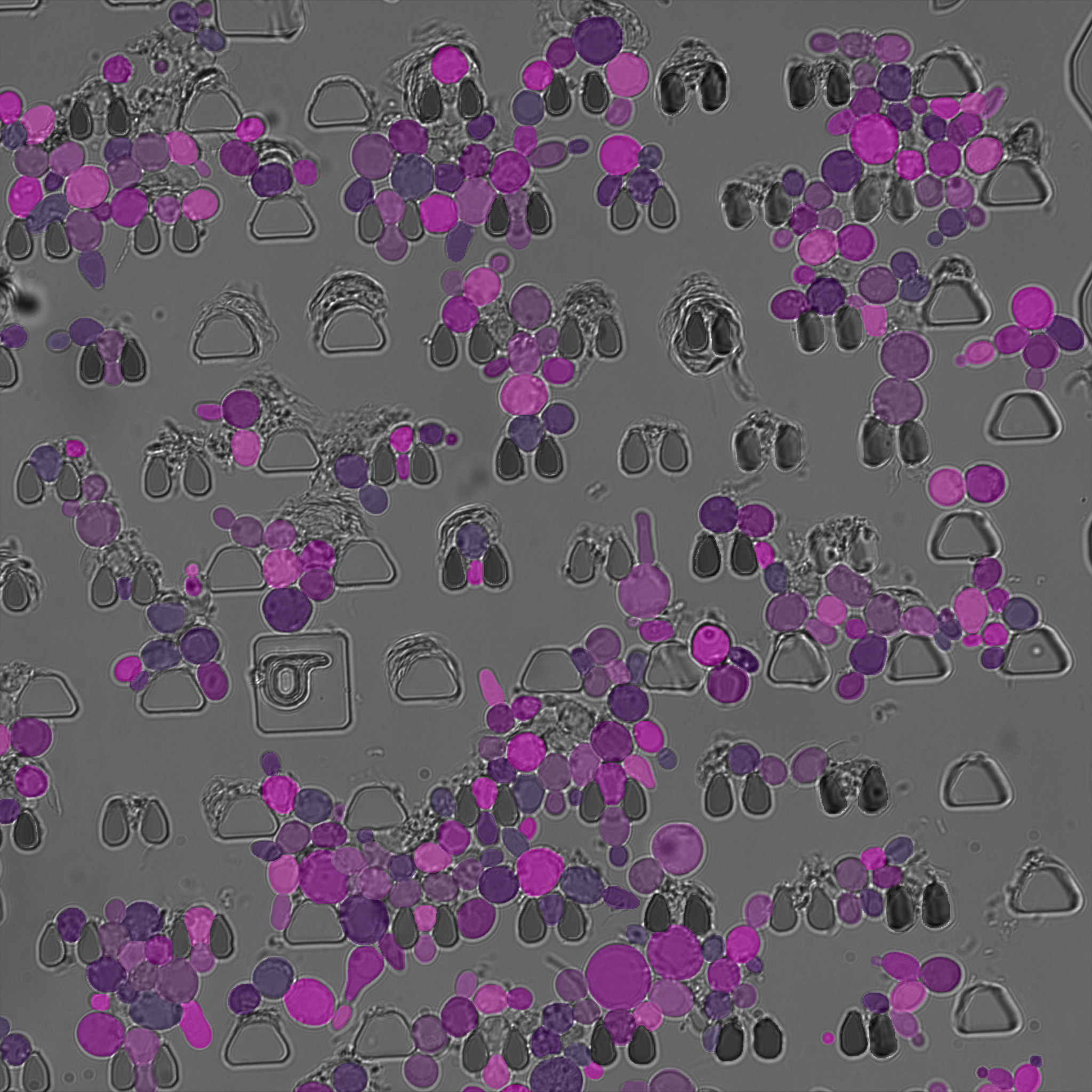}
    \captionof{figure}{\textbf{Dataset teaser.} Samples of our trapped yeast cell dataset of high-resolution ($\geq2048\times2048$) brightfield images overlaid with multi-class instance segmentation labels; shades of (dark) grey (\colorindicator{trap1}\colorindicator{trap2}\colorindicator{trap3}) indicate microstructures (traps) and shades of pink (\colorindicator{cell1}\colorindicator{cell4}\colorindicator{cell7}) indicate individual cell instances. Image brightness adjusted for better visualization. Best viewed in color; zoom in for details.\label{fig:first}\\}
    \vspace{-0.85em}
\end{center}
}]
}

\ificcvfinal\thispagestyle{empty}\fi

\begin{abstract}
    Segmenting cells and tracking their motion over time is a common task in biomedical applications. However, predicting accurate instance-wise segmentation and cell motions from microscopy imagery remains a challenging task. Using microstructured environments for analyzing single cells in a constant flow of media adds additional complexity. While large-scale labeled microscopy datasets are available, we are not aware of any large-scale dataset, including both cells and microstructures. In this paper, we introduce the trapped yeast cell (TYC) dataset, a novel dataset for understanding instance-level semantics and motions of cells in microstructures. We release $105$ dense annotated high-resolution brightfield microscopy images, including about $19\si{\kilo\relax}$ instance masks. We also release $261$ curated video clips composed of $1293$ high-resolution microscopy images to facilitate unsupervised understanding of cell motions and morphology. TYC offers ten times more instance annotations than the previously largest dataset, including cells and microstructures. Our effort also exceeds previous attempts in terms of microstructure variability, resolution, complexity, and capturing device (microscopy) variability. We facilitate a unified comparison on our novel dataset by introducing a standardized evaluation strategy. TYC and evaluation code are publicly available under CC BY 4.0 license.
\end{abstract}
\thispagestyle{firststyle}

\input{content/introduction}
\input{content/related_work}
\input{content/dataset}
\input{content/evaluation}
\input{content/results}
\input{content/conclusion}

\newpage

{\small
\paragraph{Acknowledgements.} We thank Bastian Alt for insightful feedback, Klaus-Dieter Voss for aid with the microfluidics fabrication, Markus Baier for help with the data hosting, and Aigerim Khairullina for contributing to data labeling. This work was supported by the Landesoffensive f{\"u}r wissenschaftliche Exzellenz as part of CompuGene (LOEWE). H.K. acknowledges support from the European Research Council (ERC) with the consolidator grant CONSYN (nr. 773196). C.R. acknowledges support by NEC Labs America, Inc.
}

{\small
\bibliographystyle{ieee_fullname}
\bibliography{references}
}

\end{document}

%% file: packages.tex
\usepackage{graphicx}
\usepackage{amsmath}
\usepackage{amsfonts}
\usepackage{amssymb}
\usepackage{amsthm}
\usepackage{booktabs}
\usepackage{pifont}
\usepackage{tabularx}
\usepackage{colortbl}
\usepackage{calc}
\usepackage{multirow}
\usepackage{makecell}
\usepackage{tikz}
\usepackage[export]{adjustbox}
\usepackage{pgfplots}
\usepackage{fontawesome}
\usepackage[per-mode=symbol, group-digits=false, detect-all=true,input-symbols={()}]{siunitx}
\usepackage{float}
\usepackage{listings}
\usepackage[inline]{enumitem}
\usepackage{algorithm}
\usepackage[format=plain,labelformat=simple,labelsep=period,font=small,skip=4pt,compatibility=false]{caption}
\usepackage[font=footnotesize,skip=2pt,subrefformat=parens]{subcaption}
\usetikzlibrary{spy}
\usetikzlibrary{shapes.arrows}
\usetikzlibrary{arrows.meta, arrows}
\usetikzlibrary{patterns,shapes.arrows}
\usetikzlibrary{positioning}
\usetikzlibrary{matrix}
\usetikzlibrary{pgfplots.groupplots}
\usepgfplotslibrary{groupplots,dateplot}
\usepgfplotslibrary{fillbetween}
\tikzset{every picture/.style={/utils/exec={\sffamily}}}
\pgfplotsset{compat=1.10}
\DeclareUnicodeCharacter{2212}{−}


%% file: colors.tex
\definecolor{tud0d}{cmyk/RGB/HTML}{0,0,0,.8/83,83,83/535353}
\definecolor{tud0c}{cmyk/RGB/HTML}{0,0,0,.6/137,137,137/898989}
\definecolor{tud0b}{cmyk/RGB/HTML}{0,0,0,.4/181,181,181/B5B5B5}
\definecolor{tud0a}{cmyk/RGB/HTML}{0,0,0,.2/220,220,220/DCDCDC}
\definecolor{tud1a}{cmyk/RGB/HTML}{.7,.4,0,0/93,133,195/5D85C3}
\definecolor{tud2a}{cmyk/RGB/HTML}{0.8,.2,0,0/0,156,218/009CDA}
\definecolor{tud3a}{cmyk/RGB/HTML}{0.7,0,.5,0/80,182,149/50B695}
\definecolor{tud4a}{cmyk/RGB/HTML}{.4,0,.8,0/175,204,80/AFCC50}
\definecolor{tud5a}{cmyk/RGB/HTML}{.2,0,.8,0/221,223,72/DDDF48}
\definecolor{tud6a}{cmyk/RGB/HTML}{0,.1,.7,0/255,224,92/FFE05C}
\definecolor{tud7a}{cmyk/RGB/HTML}{0,.3,.8,0/248,186,60/F8BA3C}
\definecolor{tud8a}{cmyk/RGB/HTML}{0,.6,.8,0 /238,122,52/EE7A34}
\definecolor{tud9a}{cmyk/RGB/HTML}{0,.8,.7,0/233,80,62/E9503E}
\definecolor{tud10a}{cmyk/RGB/HTML}{.2,.9,0,0/201,48,142/C9308E}
\definecolor{tud11a}{cmyk/RGB/HTML}{.6,.8,0,0/128,69,151/804597}
\definecolor{tud1b}{cmyk/RGB/HTML}{1,.6,0,0/0,90,169/005AA9}
\definecolor{tud2b}{cmyk/RGB/HTML}{1,.3,0,0/0,131,204/0083CC}
\definecolor{tud3b}{cmyk/RGB/HTML}{1,0,.6,0/0,157,129/009D81}
\definecolor{tud4b}{cmyk/RGB/HTML}{.5,0,1,0/153,192,0/99C000}
\definecolor{tud5b}{cmyk/RGB/HTML}{.3,0,1,0/201,212,0/C9D400}
\definecolor{tud6b}{cmyk/RGB/HTML}{0,.2,1,0/253,202,0/FDCA00}
\definecolor{tud7b}{cmyk/RGB/HTML}{0,.4,1,0/245,163,0/F5A300}
\definecolor{tud8b}{cmyk/RGB/HTML}{0,.7,1,0/236,101,0/EC6500}
\definecolor{tud9b}{cmyk/RGB/HTML}{0,1,.9,0/230,0,26/E6001A}
\definecolor{tud10b}{cmyk/RGB/HTML}{.4,1,0,0/166,0,132/A60084}
\definecolor{tud11b}{cmyk/RGB/HTML}{.7,1,0,0/114,16,133/721085}
\definecolor{tud1c}{cmyk/RGB/HTML}{1,.7,.2,0/0,78,138/004E8A}
\definecolor{tud2c}{cmyk/RGB/HTML}{1,.5,.2,0/0,104,157/00689D}
\definecolor{tud3c}{cmyk/RGB/HTML}{1,.2,.6,0/0,136,119/008877}
\definecolor{tud4c}{cmyk/RGB/HTML}{.6,.1,1,0/127,171,22/7FAB16}
\definecolor{tud5c}{cmyk/RGB/HTML}{.4,.1,1,0/177,189,0/B1BD00}
\definecolor{tud6c}{cmyk/RGB/HTML}{.2,.3,1,0/215,172,0/D7AC00}
\definecolor{tud7c}{cmyk/RGB/HTML}{.2,.5,1,0/210,135,0/D28700}
\definecolor{tud8c}{cmyk/RGB/HTML}{.2,.8,1,0/204,76,3/CC4C03}
\definecolor{tud9c}{cmyk/RGB/HTML}{.3,1,.9,0/185,15,34/B90F22}
\definecolor{tud10c}{cmyk/RGB/HTML}{.5,1,.3,0/149,17,105/951169}
\definecolor{tud11c}{cmyk/RGB/HTML}{.8,1,.2,0/97,28,115/611C73}
\definecolor{tud1d}{cmyk/RGB/HTML}{1,.9,.3,0/36,53,114/243572}
\definecolor{tud2d}{cmyk/RGB/HTML}{1,.7,.4,0/0,78,115/004E73}
\definecolor{tud3d}{cmyk/RGB/HTML}{1,.4,.7,0/0,113,94/00715E}
\definecolor{tud4d}{cmyk/RGB/HTML}{.7,.3,1,0/106,139,55/6A8B22}
\definecolor{tud5d}{cmyk/RGB/HTML}{.5,.2,1,0/153,166,4/99A604}
\definecolor{tud6d}{cmyk/RGB/HTML}{.4,.4,1,0/174,142,0/AE8E00}
\definecolor{tud7d}{cmyk/RGB/HTML}{.3,.6,1,0/190,111,0/BE6F00}
\definecolor{tud8d}{cmyk/RGB/HTML}{.4,.8,1,0/169,73,19/A94913}
\definecolor{tud9d}{cmyk/RGB/HTML}{.5,1,.9,0/156,28,38/961C26}
\definecolor{tud10d}{cmyk/RGB/HTML}{.7,1,.5,0/115,32,84/732054}
\definecolor{tud11d}{cmyk/RGB/HTML}{.9,1,.3,0/76,34,106/4C226A}

\definecolor{necgray}{HTML}{7d7d7d}
\definecolor{necblue}{HTML}{2c328f}
\definecolor{necbluedark}{HTML}{00285e}
\definecolor{necorange}{HTML}{eb6e00}
\definecolor{necyellow}{HTML}{fff484}
\definecolor{green}{HTML}{388a73}

\definecolor{qp51}{rgb}{0.5,0,0}
\definecolor{qp0}{rgb}{0,0,0.5}

\definecolor{tud3b6b1}{RGB}{63, 168, 97}
\definecolor{tud3b6b2}{RGB}{127, 180, 65}
\definecolor{tud3b6b3}{RGB}{190, 191, 32}

\definecolor{cell}{rgb}{1., 0., 0.89019608}
\definecolor{trap}{rgb}{0.25, 0.25, 0.25}

\definecolor{trap1}{rgb}{0.05, 0.05, 0.05}
\definecolor{trap2}{rgb}{0.2, 0.2, 0.2}
\definecolor{trap3}{rgb}{0.3, 0.3, 0.3}

\definecolor{cell1}{rgb}{1., 0., 0.89019608}
\definecolor{cell2}{rgb}{0.3, 0., 0.45}
\definecolor{cell3}{rgb}{0.5, 0., 0.53333333}
\definecolor{cell4}{rgb}{0.7, 0., 0.70980392}
\definecolor{cell5}{rgb}{1., 0.25, 0.89019608}
\definecolor{cell6}{rgb}{0.7, 0.25, 0.70980392}
\definecolor{cell7}{rgb}{0.5, 0.2, 0.55294118}
\definecolor{cell8}{rgb}{0.3, 0.1, 0.45}

%% file: content/introduction.tex
{\begin{table*}[ht!]
    \centering
    \input{tables/dataset_comparison}
    \caption{\textbf{Dataset comparison.} High-level statistics of our TYC dataset (labeled set) and other segmentation dataset including cells and microstructures. Our TYC datasets exceeds existing efforts in all high-level statistics.}
    \label{tab:dataset_comparison}
    \vspace{-0.5em}
\end{table*}}

\section{Introduction} \label{sec:introduction}

Detecting, segmenting, and tracking individual cells in microscopy images is fundamental for many biomedical applications~\cite{Emami2021, Mavska2023, Meijering2012, Ulman2017}. For example, accurate segmentation and tracking of individual cells are essential for analyzing the cellular processes of living cells in time-lapse fluorescence microscopy (TLFM) experiments~\cite{Bakker2018, Prangemeier2022}. In general, computer vision-based single-cell analysis can aid the development of personalized medicine, early tumor detection, and the analysis of biological signal transduction, amongst others~\cite{Jeckel2021, Leygeber2019, Prangemeier2020b, Prangemeier2022, Sun2020}.

Deep neural networks have become the predominant workhorse of current state-of-the-art algorithms in the domain of computer vision~\cite{Cordts2016, Lecun2015, Russakovsky2015, Voulodimos2018}. However, unleashing the full potential of deep neural networks requires vast amounts of data~\cite{Brown2020, Kirillov2023, Zhai2022}. Subsequently, the widespread availability of large-scale public datasets, such as ImageNet~\cite{Russakovsky2015}, KITTI~\cite{Geiger2013}, Cityscapes~\cite{Cordts2016}, Ego4D~\cite{Grauman2022}, SA-1B~\cite{Kirillov2023}, and LAION-5B~\cite{Schuhmann2022}, has been a major contributor to the recent success of deep neural network approaches. While large-scale (labeled) datasets in the domain of single-cell analysis are available, these datasets do not consider microstructured environments~\cite{Caicedo2019, Edlund2021, Parekh2019}. Significantly limiting the applicability of models trained without microstructures to analyze cells in microstructures due to the significant domain gap.

Microstructured environments are commonly employed to analyze populations of hundreds or thousands of cells individually~\cite{Bakker2018, Crane2014, Grunberger2014, Luo2019, Narayanamurthy2017, Pang2020, Prangemeier2020, Prangemeier2022, Sinha2022}. A constant flow of growth media hydrodynamically traps cells within the microstructures, enabling their analysis over time~\cite{Bakker2018, Prangemeier2022}. Understanding the instance-level semantics and motions of cells in microstructured environments is particularly challenging due to the perceptual similarity of microstructures and cells (\cf \cref{fig:first}). In this paper, we propose the trapped yeast cell dataset for understanding instance-level semantics and motions of trapped yeast cells. Our TYC dataset of high-resolution ($\geq2048\times2048$) brightfield microscopy images includes both a labeled instance segmentation set and an unlabeled set of video clips. In total, we release $18946$ instance masks, of cells and microstructures (\cf \cref{fig:first}). Our labeled dataset exceeds all existing cells in microstructures datasets in terms of annotated pixels, number of instances (cells \& traps), resolution, and microstructure variability (\cf \cref{tab:dataset_comparison}). Our unlabeled set contains $261$ curated high-resolution microscopy video clips for unsupervised understanding cell motions facilitating cell tracking.

In addition to the proposed dataset, we also present a standardized evaluation strategy to facilitate fair comparison of future work with our dataset. This may also serve to standardize the results of biological investigation, making them comparable between laboratories. In particular, we present an instance-level and semantic-level evaluation strategy for our labeled set. Our evaluation strategy considers both the downstream biological application and the raw segmentation performance for benchmarking.

To showcase the complexity of our TYC dataset, we report qualitative segmentation results of the Segment Anything Model (SAM), a recent foundation model for segmentation~\cite{Kirillov2023}. While SAM is able to provide fairly accurate instance masks for very simple cell and microstructure configurations, SAM fails to provide useful masks as scenes get more complex (\eg, touching cells). This demonstrates the complexity of segmenting cells in microstructures.

%% file: tables/dataset_comparison.tex
{
\small
\begin{tabular*}{\textwidth}{@{\extracolsep{\fill}}l@{}S[table-format=2.2]S[table-format=5.0]S[table-format=4.0]c@{}c@{}c@{}}
	\toprule
	Dataset & {\# annotated pixels [10\textsuperscript{7}]} & {\# cells} & {\# traps} & Resolution & Annotation type & \# trap types \\
	\midrule
	Bakker \etal~\cite{Bakker2018} & {--} & 1000 & 0 & \hphantom{$<$}512$\times$512 & Cell outline & 1 \\
	Reich \etal~\cite{Reich2023} & 0.81 & 914 & 971 & \hphantom{$<$}128$\times$128 & Instance segmentation & 2 \\
	\midrule
	\textbf{TYC dataset (ours)} & 46.39 & 14541 & 4405 & $\geq$2048$\times$2048 & Instance segmentation & 6 \\
	\bottomrule
\end{tabular*}}

%% file: content/related_work.tex
{
\begin{figure*}[ht!]
    \centering
    \includegraphics[width=\textwidth]{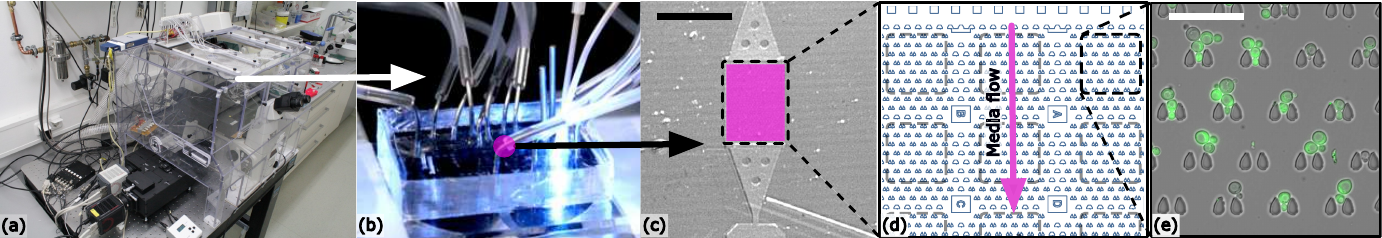}
    \caption{\textbf{Overview data acquisition process.} General TLFM experiment setup for single-cell fluorescence measurement. Microscope (from left to right), microfluidic chip (cell chamber within the purple dot), trap chamber (pink rectangle), trap microstructure design, traps with cells (cells in green, fluorescent overlay, right); black scale bar \SI{1}{\milli\meter} white scale bar \SI{10}{\micro\meter}.}
    \label{fig:data_acquisition}
    \vspace{-0.5em}
\end{figure*}
}

\section{Related Work} \label{sec:related_work}

Many biological applications rely on microscopy image data for single-cell analysis~\cite{Merchant2022}. Most applications require instance segmentation of cells~\cite{Meijering2012, Meijering2016, Merchant2022}. Object tracking is required to analyze cells through time~\cite {Mavska2023}. The vast majority of state-of-the-art cell segmentation approaches rely on deep neural networks~\cite{Moen2019, Falk2019}; prominent examples include, U-Net~\cite{Ronneberger2015} StarDist~\cite{Schmidt2018}, and Cellpose~\cite{Pachitariu2022, Stringer2021}. Similarly, current cell tracking approaches also predominately rely on deep neural networks~\cite{Ben2022, Chang2017, Jang2023, Loffler2022, Lugagne2020, Mavska2023, Ulman2017, Scherr2020}.

The widespread use of deep learning in cell analytics tasks has motivated the construction of large and general datasets~\cite {Moen2019, Edlund2021}. Caicedo~\etal~\cite{Caicedo2019} proposed the 2018 Data Science Bowl dataset for single-cell segmentation in fluorescence imagery, composed of $670$ images with instance-wise annotations. Recently, Edlund~\etal~\cite{Edlund2021} released the large-scale microscopy image dataset  LIVECell, for single-cell (instance) segmentation, with $5239$ densely annotated phase-contrast images and eight cell types. The ISBI Cell Tracking Challenge~\cite{Mavska2023, Ulman2017} offers multiple datasets for both cell segmentation and tracking. Anjum~\etal~\cite{Anjum2020} proposed CTMC with $86$ annotated videos.

While significant efforts have been devoted to the development of cell analytics datasets, only very limited (annotated) data of cells in microstructures is available. Bakker~\etal~\cite{Bakker2018} released about $1000$ cell outline annotations of yeast cells in microstructures. Recently, Reich~\etal~\cite{Reich2023} proposed the first instance segmentation dataset of yeast cells in microstructured environments. These datasets, however, entail major limitations (\cf \cref{tab:dataset_comparison}). Both datasets include only a limited number of annotated objects (cells and traps) and capture only a small number of different trap types. Additionally, the dataset by Reich~\etal~\cite{Reich2023} only includes low-resolution crops of high-resolution brightfield microscopy images. The annotated set of our TYC dataset exceeds all previous datasets of cells in microstructures in terms of all high-level dataset statistics. \cref{tab:dataset_comparison} provides a comprehensive overview between existing datasets including cells and microstructures and our TYC dataset (labeled set).

To overcome the limited availability of annotated microscopy data, some approaches aim to synthesize microscopy images and annotations~\cite{Kruitbosch2022, Sachs2022}. While synthesized data can be effective for simple cell analytic tasks, these approaches are not able to synthesize complicated cell and microstructure scenes as well as vast and complex cell motions. Additionally, synthetic images commonly suffer from a domain gap \wrt real images, requiring approaches to address this domain gap when deep neural networks are pre-trained on synthetic data~\cite{Farahani2021, Nikolenko2021, Wang2018}.

%% file: content/dataset.tex
\section{Dataset} \label{sec:dataset}

We made numerous design choices during data collection, curation, and annotation of our TYC dataset. This section reports these design decisions and provides dataset statistics. We selected $105$ high-resolution ($\geq2048\times2048$) brightfield images for our labeled set, from over a dozen TLFM experiments. The unlabelled video set contains $261$ manually-selected video clips. The selection encompasses the entirety of the data distribution: it includes edge cases, various magnifications, and diverse focal positions. The key objects in the imagery are yeast cells (various strains) and trap microstructures. Six widespread trap geometries are included in the dataset (\cf \cref{fig:trap_types}), in order to ensure that models trained on our TYC dataset can generalize over a wide variety of microstructure geometries.

\subsection{Data acquisition} \label{subsec:data_acquisition}

\begin{figure}[t]
    \centering
    \begin{subfigure}{.158\linewidth}
        \centering
        \includegraphics[width=\linewidth, frame]{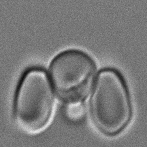}\\[0.66pt]
        \includegraphics[width=\linewidth, frame]{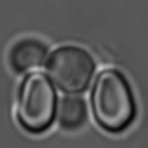}
        \caption*{Type 1}
        \label{subfig:trap_type_1}
    \end{subfigure}\hfill%
    \begin{subfigure}{.158\linewidth}
        \centering
        \includegraphics[width=\linewidth, frame]{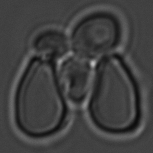}\\[0.66pt]
        \includegraphics[width=\linewidth, frame]{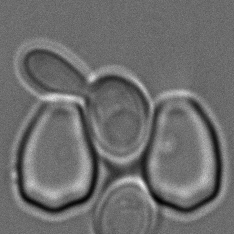}
        \caption*{Type 2}
        \label{subfig:trap_type_2}
    \end{subfigure}\hfill%
    \begin{subfigure}{.158\linewidth}
        \centering
        \includegraphics[width=\linewidth, height=\linewidth, frame]{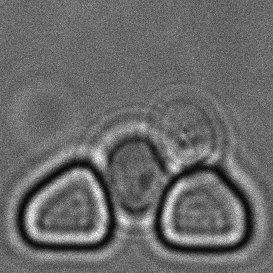}\\[0.75pt]
        \includegraphics[width=\linewidth, height=\linewidth, frame]{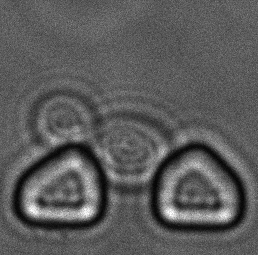}
        \caption*{Type 3}
        \label{subfig:trap_type_3}
    \end{subfigure}\hfill%
    \begin{subfigure}{.158\linewidth}
        \centering
        \includegraphics[width=\linewidth, frame]{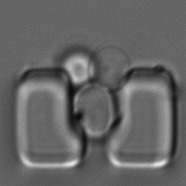}\\[0.66pt]
        \includegraphics[width=\linewidth, frame]{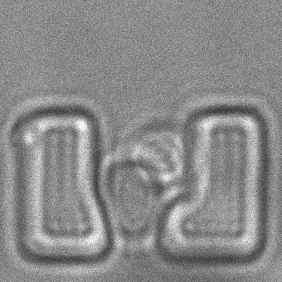}
        \caption*{Type 4}
        \label{subfig:trap_type_4}
    \end{subfigure}\hfill%
    \begin{subfigure}{.158\linewidth}
        \centering
        \includegraphics[width=\linewidth, frame]{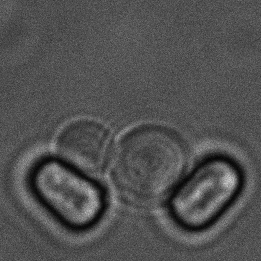}\\[0.66pt]
        \includegraphics[width=\linewidth, frame]{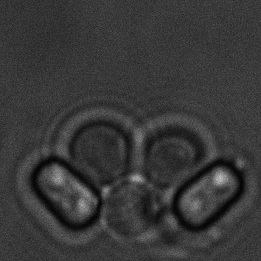}
        \caption*{Type 5}
        \label{subfig:trap_type_5}
    \end{subfigure}\hfill%
    \begin{subfigure}{.158\linewidth}
        \centering
        \includegraphics[width=\linewidth, frame]{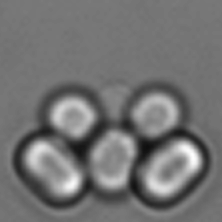}\\[0.66pt]
        \includegraphics[width=\linewidth, frame]{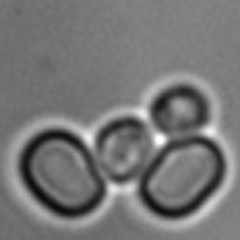}
        \caption*{Type 6}
        \label{subfig:trap_type_6}
    \end{subfigure}\hfill%
    \caption{\textbf{Trap types.} Overview of trap microstructure geometries included in our TYC dataset, cropped from high-resolution microscopy images (\cf \cref{fig:first}).}
    \label{fig:trap_types}
    \vspace{-0.65em}
\end{figure}

We recorded the dataset with a computer-controlled microscope (Nikon Eclipse Ti, \cref{fig:data_acquisition}a). The yeast cells were cultured in a tightly controlled environment within a microfluidic chip (\cref{fig:data_acquisition}b) comprised of a microscopy cover slip and transparent Polydimethylsiloxane (PDMS). The trap chamber contains approximately 1000 trap microstructures (\cf \cref{fig:data_acquisition}c). Various trap geometries are available (\cf \cref{fig:trap_types}). The microstructures and a constant flow of yeast growth media hydrodynamically trap the cells, constraining them laterally in XY (\cf \cref{fig:data_acquisition}d \& \ref{fig:data_acquisition}e). The axial constraint is provided by the coverslip and PDMS ceiling, the space between which is on the order of a cell diameter. This facilitates a continuously uniform focus of the cells. The entire chip is maintained at a temperature of $30$ °C and together with the flow of yeast growth media this enables yeast to grow for prolonged periods and over multiple cell cycles. The cells bud daughter cells multiple times within the course of a typical experiment (budding approximately every $90\si{\minute}$, experiments typically run for $15\si{\hour}$), and these are flushed out by the continuous flow to avoid clogging of the chip. 

We recorded time-lapse brightfield (transmitted light) and fluorescent channel imagery of the budding yeast cells every $10\si{\minute}$. We used either NIS-Elements or $\mu$Manager to control the microscope in different experiments~\cite{Edelstein2010}. Imagery from both $100\times$ and $60\times$ objectives, which are widespread in order to resolve the cells, are included. A CoolLED pE-$100$ and a Lumencor SpectraX light engine illuminated the respective channels and lighting conditions are varied throughout the dataset. Resolutions are $2048\times2048$ and $2304\times2304$ recorded with Hamamatsu cameras ORCA-Flash4.0 V2, V3, and with an ORCA-Fusion. Multiple lateral and axial positions were recorded sequentially at each timestep (\cref{fig:data_acquisition}). In addition to our own data, selected imagery from the dataset by Bakker~\etal~\cite{Bakker2018, Crane2014} are included in our TYC dataset (upscaled from $512\times512$ to $2048\times2048$ to match our data).

\subsection{Biological application} \label{subsec:data_application}

Creating novel functionality from standardized parts is central to synthetic biology. Ideally, synthetic biological circuitry is rationally designed \textit{in silico} to predetermined design specifications \cite{Bittihn2018, Prangemeier2020}, \eg, to detect and kill cancer \cite{Si2018, Xie2011}, but also for a plethora of other applications and real-world problems. \textit{Saccharomyces cerevisiae} (yeast) and \textit{E. coli} are the model organisms of choice in synthetic biology \cite{Prangemeier2020}, both in terms of development and characterization of novel biological circuitry. With TYC we focus on yeast.

TLFM plays a central role in the thorough characterization and standardization of biological investigation, as it is the only technique available to capture both the population heterogeneity and dynamics of synthetic circuitry on the single-cell level \cite{Leygeber2019, Locke2009, Longo2006}. However, extracting the relevant information from within the imagery, \ie measuring fluorescence via cell segmentation, is currently a limiting factor and there is no standard for it or calibration set to compare vision model-based measurements upon. The TYC dataset will not only facilitate the development of new vision models that resolve the current bottleneck in extracting the measurement data reliably but also provide a benchmark for all of the future models to be calibrated against and contribute to the reproducibility of TLFM-based biological investigation and development.

\subsection{Annotations and classes} \label{subsec:data_classes}

\begin{figure}[t]
    \centering
    \begin{subfigure}{.49\linewidth}
        \centering
        \includegraphics[width=\linewidth, trim={0 0cm 0 29cm}, clip, frame, decodearray={0.05 1.0 0.05 1.0 0.05 1.0}]{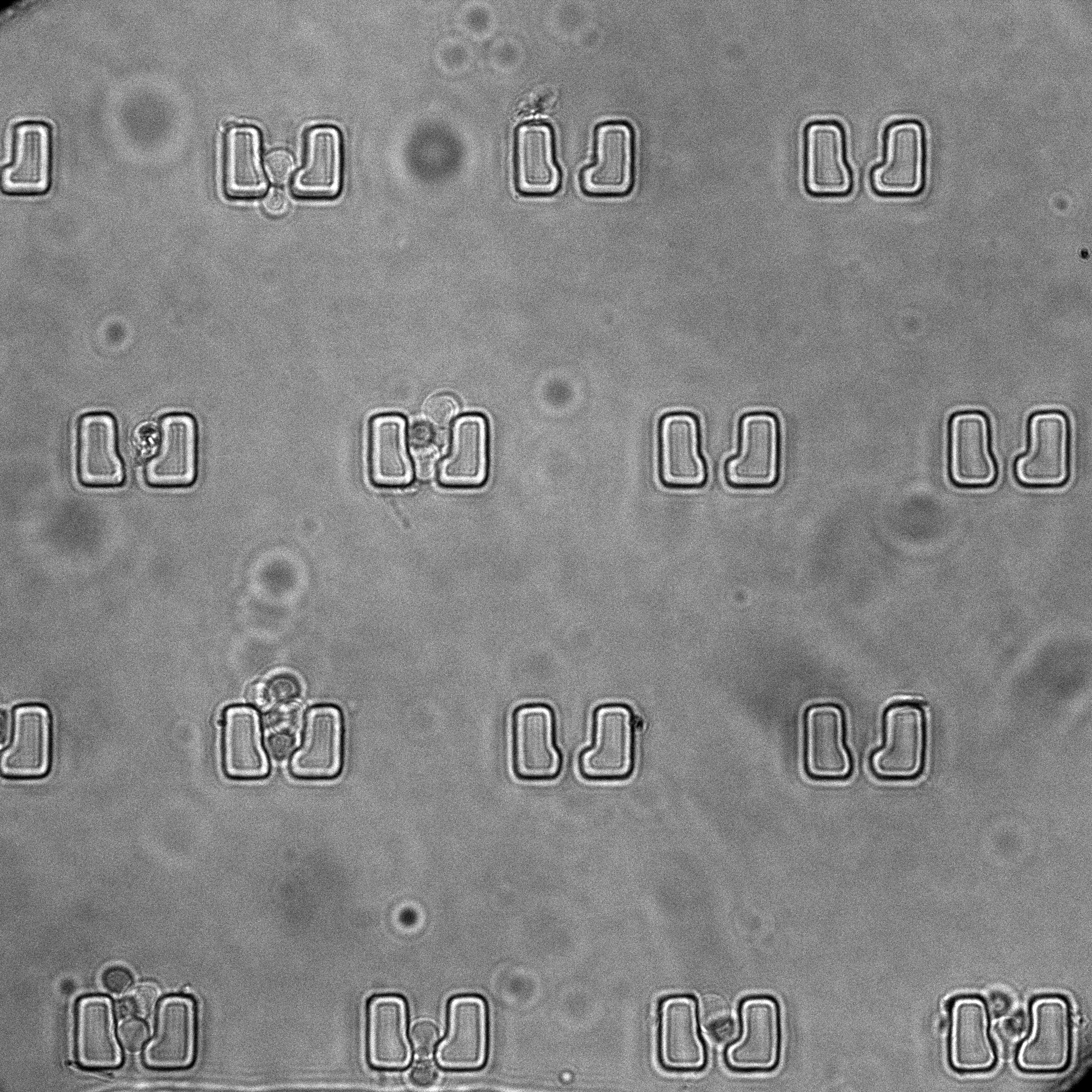}\\[3pt]
        \includegraphics[width=\linewidth, trim={0 0cm 0 29cm}, clip, frame, decodearray={0.05 1.0 0.05 1.0 0.05 1.0}]{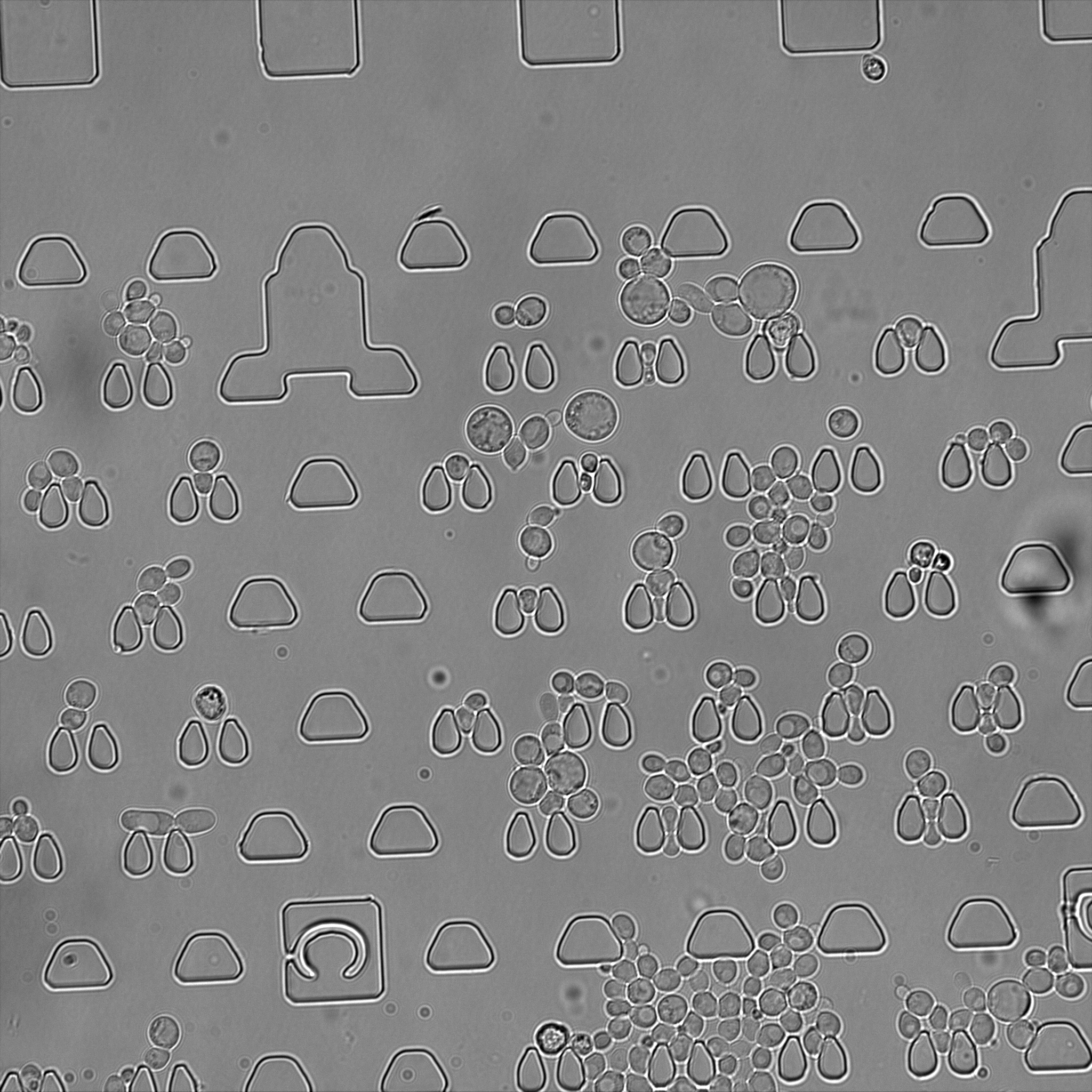}
        \caption{Brightfield microscopy image}
        \label{subfig:image}
    \end{subfigure}\hfill%
    \begin{subfigure}{.49\linewidth}
        \centering
        \includegraphics[width=\linewidth, frame, trim={0 0cm 0 29cm}, clip]{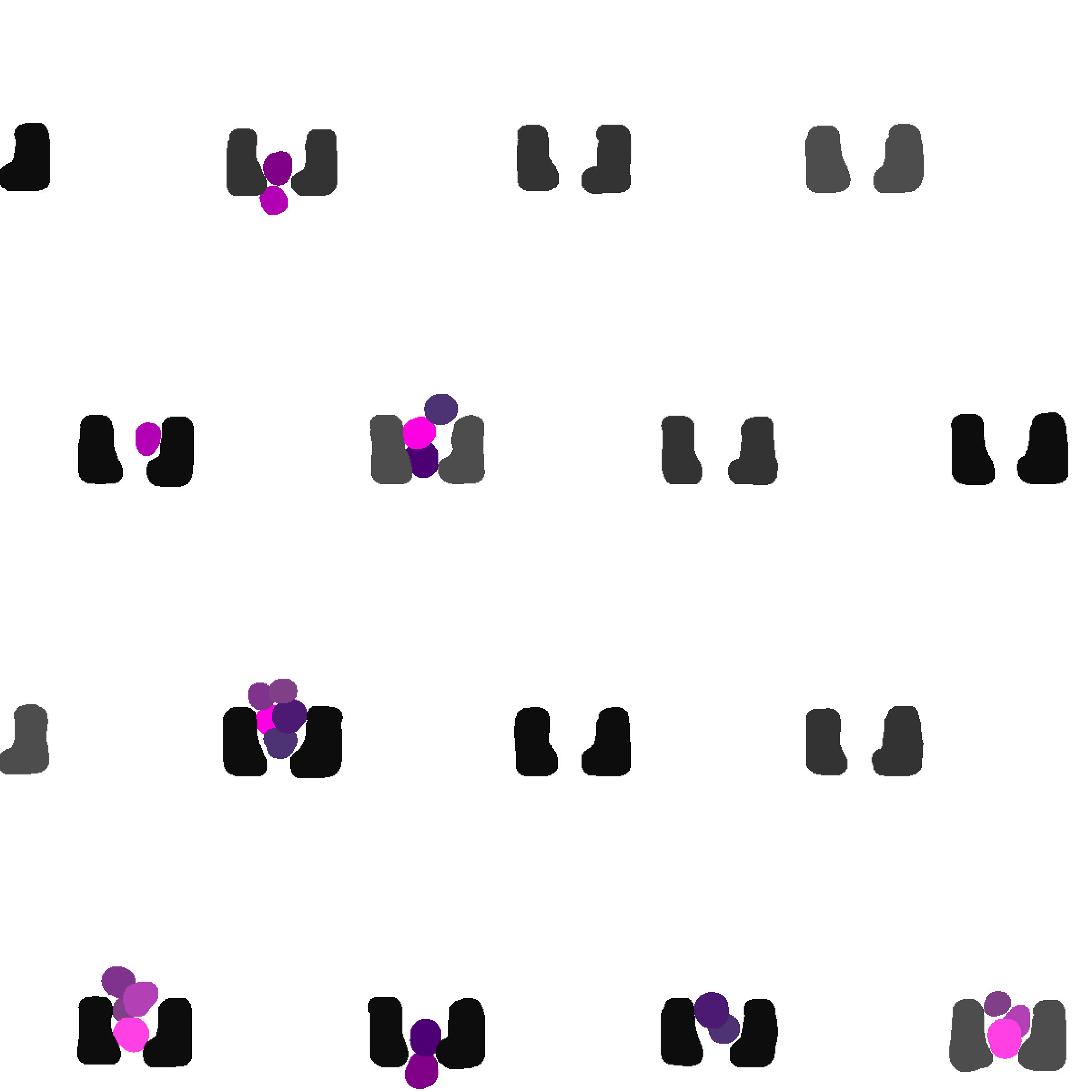}\\[3pt]
        \includegraphics[width=\linewidth, frame, trim={0 0cm 0 29cm}, clip]{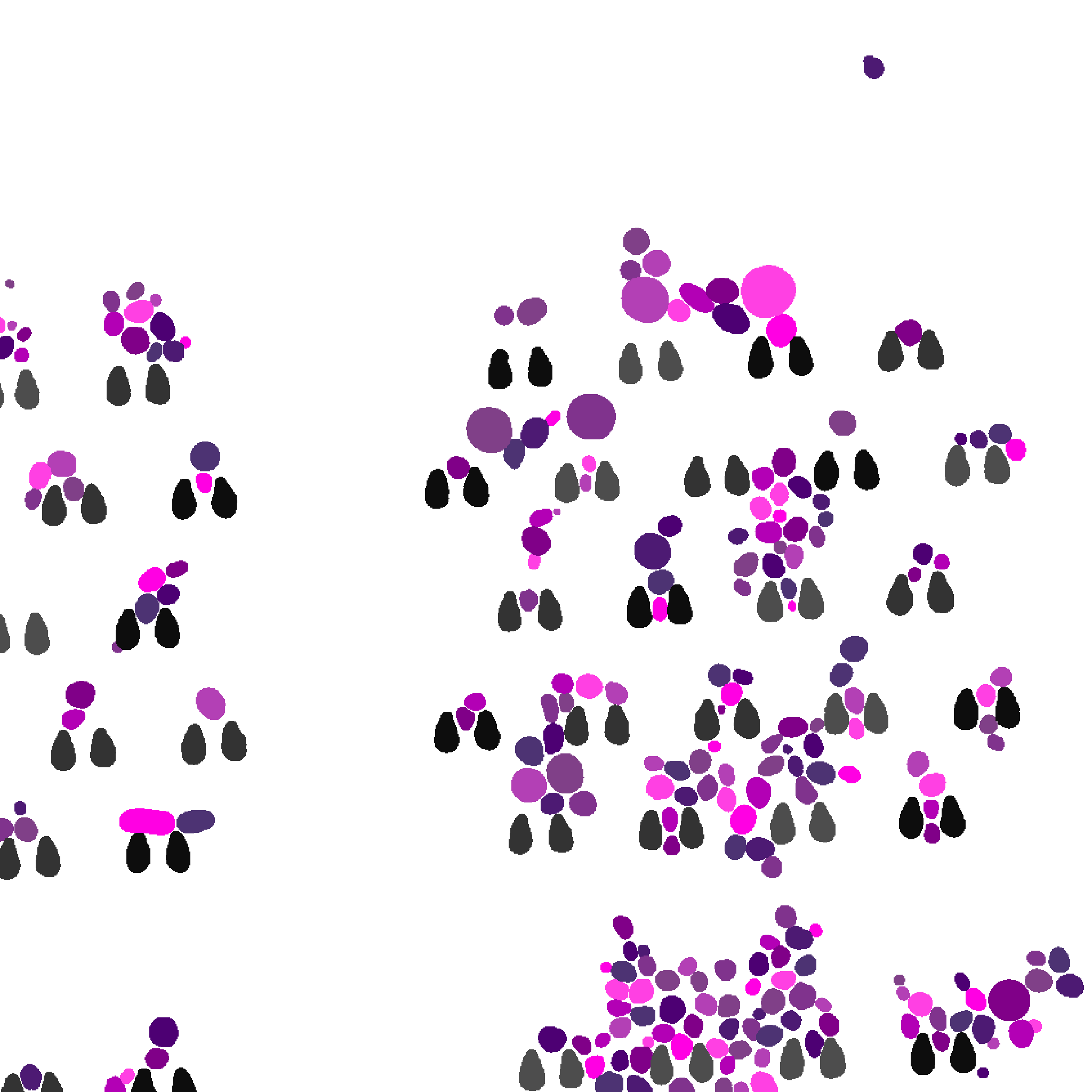}
        \caption{Instance segmentation label}
        \label{subfig:label}
    \end{subfigure}\hfill%
    \caption{\textbf{Instance annotations.} Brightfield microscopy images including cells \& microstructures on the left \emph{\subref{subfig:image}} and the corresponding instance segmentation labels on the right \emph{\subref{subfig:label}}. Shades of grey (\colorindicator{trap1}\colorindicator{trap2}\colorindicator{trap3}) indicate microstructures (traps) and shades of pink (\colorindicator{cell1}\colorindicator{cell4}\colorindicator{cell7}) indicate individual cell. For visualization purposes, we adjusted the brightness of the microscopy image and show a crop of the full image. Best viewed in color; zoom in for details.}
    \label{fig:instance_annotations}
    \vspace{-0.65em}
\end{figure}

We present a labeled set composed of $105$ high-resolution brightfield microscopy images with instance-level annotations for cells and microstructures. An example of our annotated microscopy images is presented in \cref{fig:instance_annotations}.

Our dense annotations consist of non-overlapping pixel-wise instance maps with semantic class annotation. We annotated all images in-house to ensure consistent and high-quality annotations. Annotation time per image differs between about $30\si{\minute}$ for microscopy images with only a few cells ($<20$ cell instance) and up to $2\si{\hour}$ for images with more than $300$ cell instances (\cf \cref{fig:first} and \ref{fig:instance_annotations}).

We assume no overlap between instances since the employed microfluidic chips ensure a monolayer of cells. For ambiguous configurations (\eg, touching and slightly overlapping cells), we label ambiguous regions to belong to the upper cell instance. Note due to accurate microfluidic chip production, ambiguous cases are a rarity.

Following prior work, our instance segmentation annotations include two semantic classes, cell and (microstructure) trap~\cite{Reich2023}. While instance segmentations of merely the cells suffice for most applications, we also include instance-level annotations of microstructures~\cite{Merchant2022}. We motivate this design choice twofold. First, learning to distinguish between perceptually similar cell and trap instances might be enforced by explicitly learning to segment both cells and traps. Second, understanding the location of both cell and trap instance enables us to determine cells that are hydrodynamically trapped and those that are not trapped, which are likely washed out of the image and the whole chip~\cite{Prangemeier2020, Reich2023}.

All pixels not labeled as a trap or cell instance are assumed to belong to the background. Since we only assume a single background class, our annotations can also be seen as panoptic segmentation labels~\cite{Kirillov2019}. In particular, panoptic segmentation distinguishes between \textit{things} - countable objects like cells or traps and \textit{stuff} - amorphous regions such as the background. We later use this property during evaluation and assume that cell and trap instances belong to the things category and the background to the stuff category.

\begin{figure}[b]
    \centering
    \vspace{-0.65em}
    \includegraphics[width=0.19\columnwidth, frame]{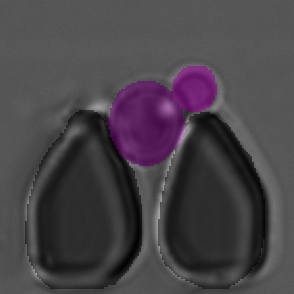}\hfill%
    \includegraphics[width=0.19\columnwidth, frame]{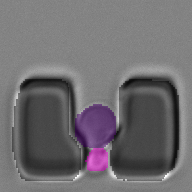}\hfill%
    \includegraphics[width=0.19\columnwidth, frame]{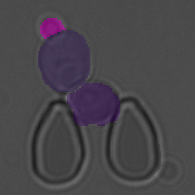}\hfill%
    \includegraphics[width=0.19\columnwidth, frame]{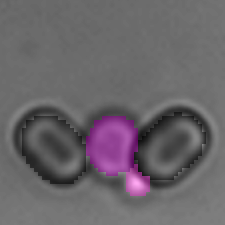}\hfill%
    \includegraphics[width=0.19\columnwidth, frame]{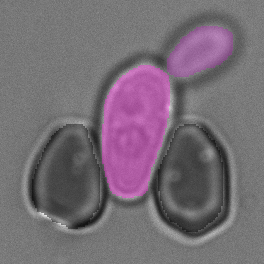}\\[1pt]
    \includegraphics[width=0.19\columnwidth, frame]{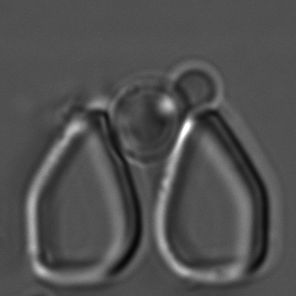}\hfill%
    \includegraphics[width=0.19\columnwidth, frame]{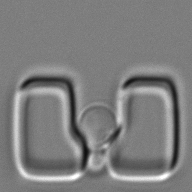}\hfill%
    \includegraphics[width=0.19\columnwidth, frame]{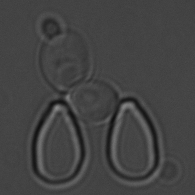}\hfill%
    \includegraphics[width=0.19\columnwidth, frame]{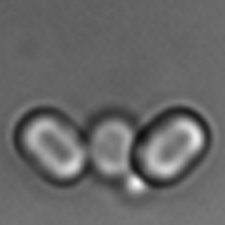}\hfill%
    \includegraphics[width=0.19\columnwidth, frame]{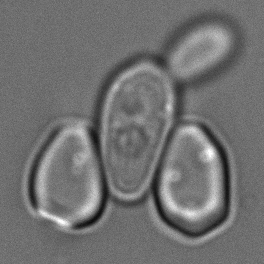}
    \caption{\textbf{Examples of budding yeast cells.} A daughter cells (small cells) bud from a trapped mother cells. The bottom row shows the raw image and the top row an overlay between the label and the raw image. Crops from full images are shown.}
    \label{fig:budding}
\end{figure}

During annotation, distinguishing between cell instances is well-defined by the cell membrane. Budding yeast cells, in which a daughter cell buds off the mother cell, represent a non-trivial edge case~\cite{Duina2014}. We label daughter cells as separate cell instances as soon as the connecting region between the mother and daughter becomes convex or thinner than the daughter cell's diameter. The process of budding is shown in \cref{fig:budding_seq}; additionally, \cref{fig:budding} provides annotated examples of budding cells.

\begin{figure}[t]
    \centering
    \includegraphics[width=0.19\columnwidth, frame]{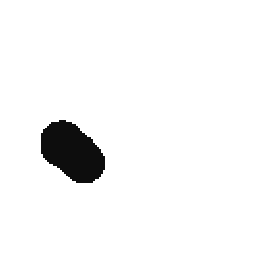}\hfill%
    \includegraphics[width=0.19\columnwidth, frame]{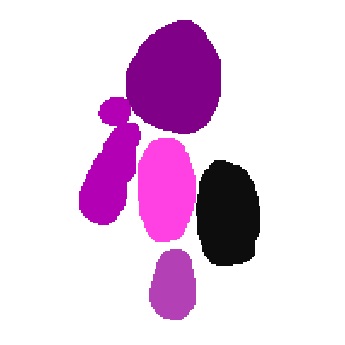}\hfill%
    \includegraphics[width=0.19\columnwidth, frame]{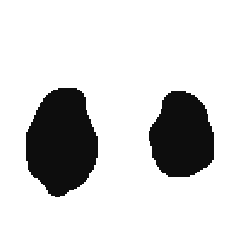}\hfill%
    \includegraphics[width=0.19\columnwidth, frame]{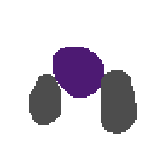}\hfill%
    \includegraphics[width=0.19\columnwidth, frame]{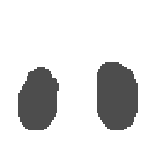}\\[1pt]
    \includegraphics[width=0.19\columnwidth, frame]{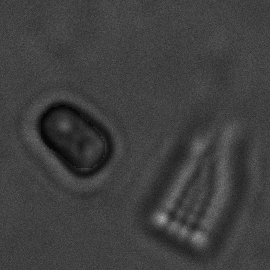}\hfill%
    \includegraphics[width=0.19\columnwidth, frame]{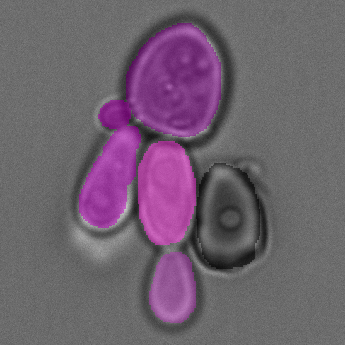}\hfill%
    \includegraphics[width=0.19\columnwidth, frame]{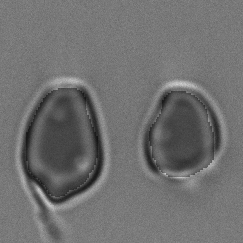}\hfill%
    \includegraphics[width=0.19\columnwidth, frame]{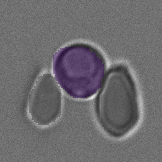}\hfill%
    \includegraphics[width=0.19\columnwidth, frame]{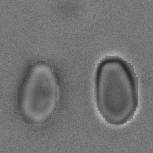}\\[1pt]
    \includegraphics[width=0.19\columnwidth, frame]{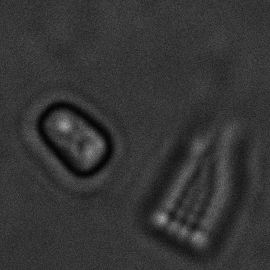}\hfill%
    \includegraphics[width=0.19\columnwidth, frame]{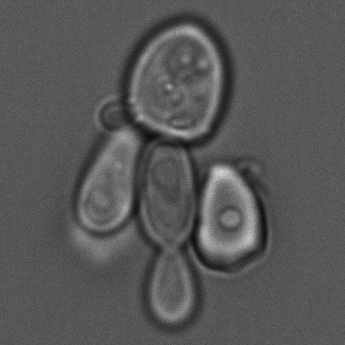}\hfill%
    \includegraphics[width=0.19\columnwidth, frame]{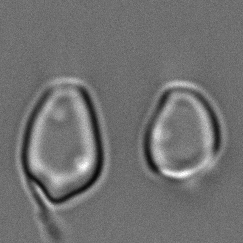}\hfill%
    \includegraphics[width=0.19\columnwidth, frame]{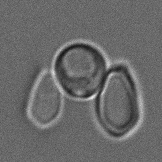}\hfill%
    \includegraphics[width=0.19\columnwidth, frame]{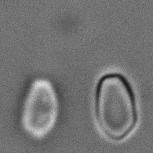}
    \caption{\textbf{Trap fabrication errors.} Examples of damaged and broken traps. The top row shows the instance segmentation label, the bottom row shows the brightfield image, and the middle row shows an overlay of both. Crops from full images are shown.}
    \label{fig:broken_trap}
    \vspace{-0.65em}
\end{figure}

Fabrication errors can lead to damaged or fully broken microstructures. While we generally label a pair of traps as a single microstructure instance (\cf \cref{fig:instance_annotations}), broken traps lead to ambiguities. If a trap is damaged, but the general structure still follows the structure of an intact trap, we annotate the full trap pair (\cf \cref{fig:broken_trap} right example). In case the structure of a broken trap is vastly different from an intact trap, we refrain from labeling the broken trap (\cf \cref{fig:broken_trap} left example). In case a trap is broken and overlapped by a cell instance, we also refrain from labeling the broken trap (\cf \cref{fig:broken_trap} second left example). Note both previous cases lead to microstructure instances, including only a single trap, not a trap pair (standard case).

\subsection{Statistical analysis} \label{subsec:data_stats}

This subsection provides a statistical analysis of our labeled TYC dataset. We also compare our dataset to the dataset proposed by Reich~\etal~\cite{Reich2023}. Note that a high-level comparison against related datasets is provided in \cref{tab:dataset_comparison}.

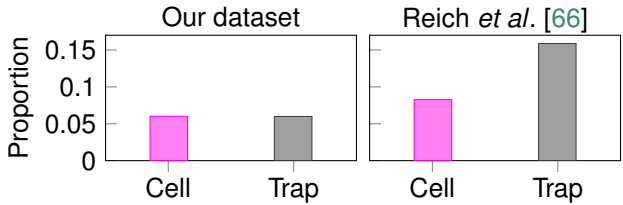
\begin{figure}[ht!]
    \centering
    \vspace{-0.5em}
    \input{artwork/semantic_distribution}
    \caption{\textbf{Semantic distribution.} Portion of labeled pixels (y-axis) per semantic class (x-axis) for our TYC dataset and \cite{Reich2023}. Note pixels not labeled as cell or trap belong to the background.}
    \label{fig:semantic_dist}
\end{figure}

\cref{fig:semantic_dist} compares the distribution of semantic annotations. $6\%$ of all annotated pixels belong to the cell class. Similarly, also $6\%$ of all annotated pixels are labeled as traps. The dataset by Reich~\etal~\cite{Reich2023} includes a higher pixel-wise density of cells ($8\%$) and traps ($16\%$). Notably, our dataset provides a more balanced semantic distribution than the dataset by Reich~\etal~\cite{Reich2023}.

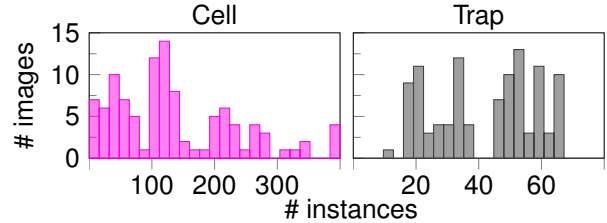
\begin{figure}[ht!]
    \centering
    \vspace{-0.35em}
    \input{artwork/instance_hist}
    \caption{\textbf{Number of instance in images}. Histogram visualizing the frequency of number of object instances (cells \& trap) in our labeled dataset. The left plot shows the cell class object counts (pink \colorindicator{cell}) and the right plot the trap class object counts (grey \colorindicator{trap!50}).}
    \label{fig:instance_hist}
    \vspace{-0.65em}
\end{figure}

Our labeled dataset of brightfield microscopy images includes vastly different numbers of object instances. On average, a microscopy image of our labeled set includes $138$ cells and $42$ trap pairs. In the extreme, the maximum number of cells in an image is $469$, and the maximum number of traps is $67$. The minimum number of cells in a microscopy image is $6$, and the minimum number of trap pair instances is $12$. \cref{fig:instance_hist} showcases a histogram of object instances per image for both the cell and trap class. It can be observed that the range of cell instance counts is significantly larger than the range of the number of trap instances.

\begin{figure}[ht!]
    \centering
    \input{artwork/object_size_hist}
    \caption{\textbf{Object size histogram.} Histogram showing the frequency of different object instance sizes (cells \& trap) in our labeled dataset. The left plot shows the cell class object sizes (pink~\colorindicator{cell}) and the right plot the trap class object sizes (grey \colorindicator{trap!50}).}
    \label{fig:size_hist}
\end{figure}
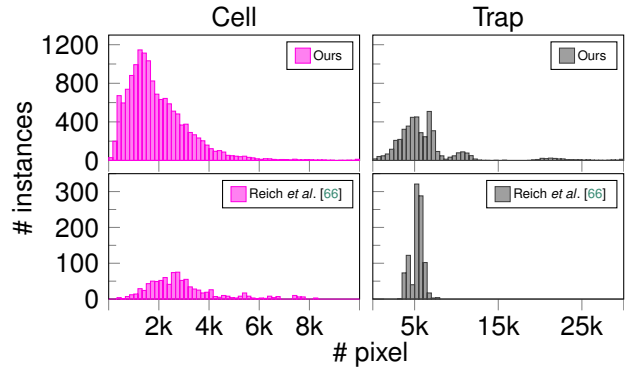

\cref{fig:size_hist} showcases the histogram of object sizes in our labeled set and the dataset by Reich~\etal~\cite{Reich2023}. We observe that while our labeled dataset includes more objects, our dataset also entails a broader distribution of cell and trap sizes. In particular, our labeled dataset includes more small cell instances than the dataset of Reich~\etal~\cite{Reich2023}. While the dataset by Reich~\etal~\cite{Reich2023} includes primary trap instances of the same shape, our dataset offers a wider range of different trap sizes.

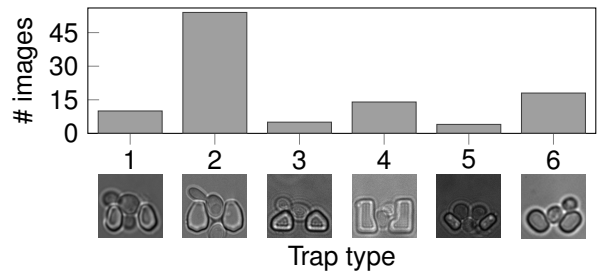
\begin{figure}[ht!]
    \centering
    \vspace{-0.5em}
    \input{artwork/trap_type_hist}
    \caption{\textbf{Frequency of trap types.} Frequency of different trap types in our TYC dataset (labeled set).}
    \label{fig:trap_type_hist}
\end{figure}

Our TYC dataset includes six different microstructure trap geometries (\cref{fig:trap_types}). The frequency of these different trap types is presented in \cref{fig:trap_type_hist}. Type 2 is routinely employed in our wet lab and is the most frequent in the dataset. Other traps serve more specialized roles.

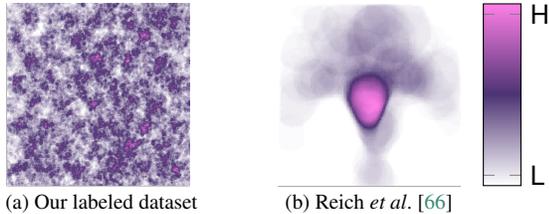
\begin{figure}[t]
    \centering
    \begin{subfigure}{.49\linewidth}
        \centering
        \input{artwork/cell_density}
        \caption{Our labeled dataset}
        \label{subfig:cell_density_ours}
    \end{subfigure}\hfill%
    \begin{subfigure}{.49\linewidth}
        \centering
        \input{artwork/cell_density_old}
        \vspace{-0.1cm}
        \caption{Reich~\etal~\cite{Reich2023}\hphantom{Christoph01}}
        \label{subfig:cell_density_old}
    \end{subfigure}\hfill%
    \caption{\textbf{Cell density map.} Cell location density map of \emph{\subref{subfig:cell_density_ours}} our dataset and \emph{\subref{subfig:cell_density_old}} \cite{Reich2023}. Areas in pink (\colorindicator{cell}) indicate a high cell density (H). Regions in white indicate a low cell density (L).}
    \label{fig:cell_density}
    \vspace{-0.65em}
\end{figure}

The spatial distribution of cells in our labeled set is visualized in \cref{fig:cell_density}. We observe that our labeled set does not have a positional bias. In contrast, the dataset by Reich~\etal~\cite{Reich2023} entails a strong positional bias, with cells primarily located in the center of the image. This bias is caused by the fact that the dataset by Reich~\etal~\cite{Reich2023} utilizes crops of the full microscopy image, including only a single trap pair.

\begin{figure}[b]
    \centering
    \vspace{-0.095cm}
    \begin{subfigure}{.49\linewidth}
        \centering
        \input{artwork/trap_density}
        \caption{Our labeled dataset}
        \label{subfig:trap_density_ours}
    \end{subfigure}\hfill%
    \begin{subfigure}{.49\linewidth}
        \centering
        \input{artwork/trap_density_old}
        \vspace{-0.095cm}
        \caption{Reich~\etal~\cite{Reich2023}\hphantom{Christoph01}}
        \label{subfig:trap_density_old}
    \end{subfigure}\hfill%
    \caption{\textbf{Trap density map.} Trap location density map of \emph{\subref{subfig:trap_density_ours}} our dataset and \emph{\subref{subfig:trap_density_old}} the dataset by Reich~\etal~\cite{Reich2023}. Black (\colorindicator{black}) indicate regions where many traps are located (H). White areas showcase regions where only a few or no traps are located (L).}
    \label{fig:trap_density}
    \vspace{-0.65em}
\end{figure}
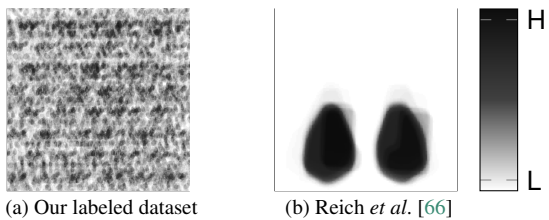

Similar to the cell positioning, the spatial distribution of traps also shows no predominant positional bias (\cf \cref{fig:trap_density}). While there is no general positional bias towards specific image regions, the regular grid-based positioning of the microstructured traps can be observed. Thus, understanding the general grid-based positioning of traps might enable accurate detection and segmentation of trap instances. In contrast to our dataset, the dataset by Reich~\etal~\cite{Reich2023} entails a strong positional bias of trap positions. This is due to the fact that Reich~\etal~\cite{Reich2023} utilizes trap-centered crops of the full microscopy images.

\begin{figure}[ht!]
    \centering
    \begin{subfigure}{.49\linewidth}
        \centering
        \includegraphics[width=\linewidth, trim={0 32cm 0 0cm}, clip, frame]{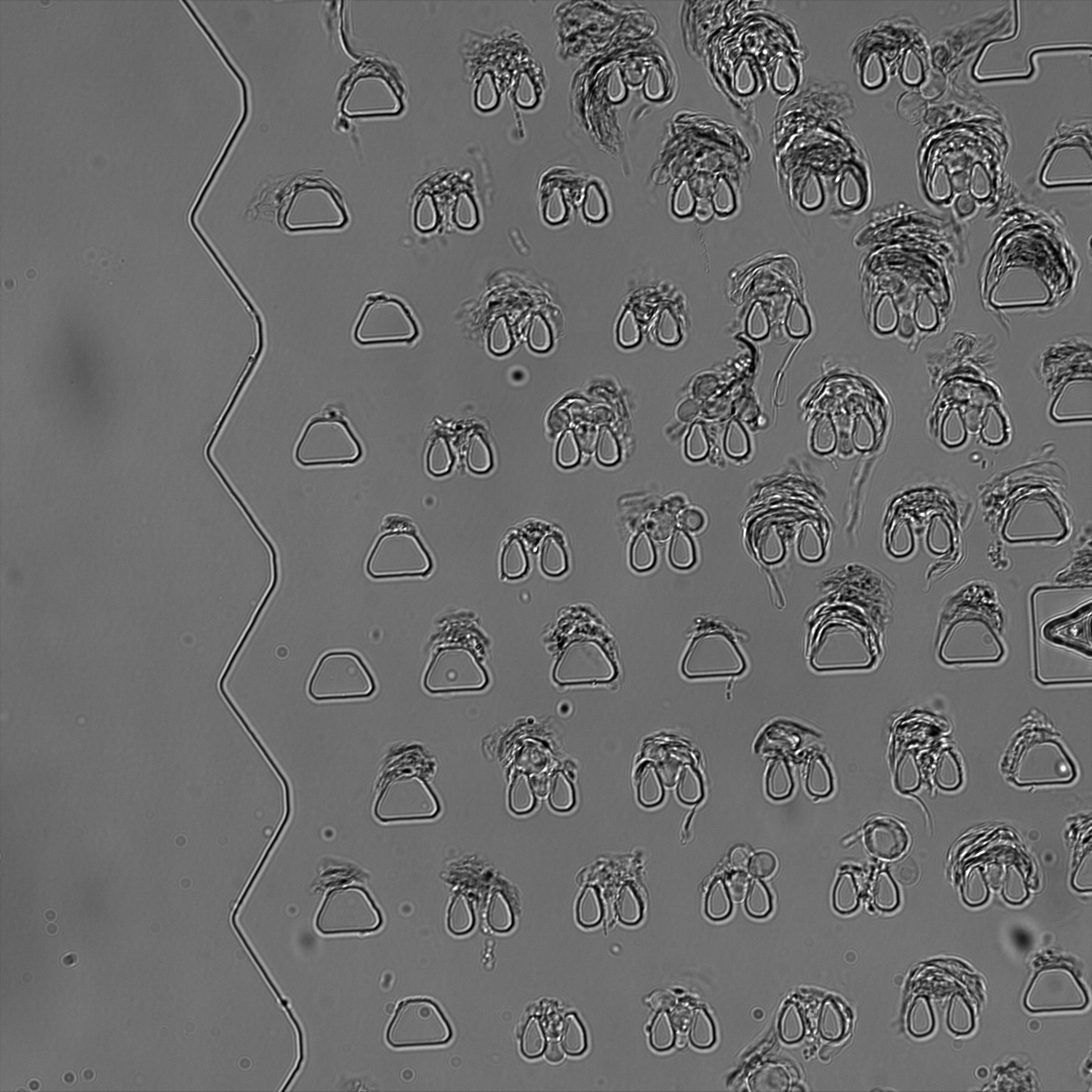}
        \caption{Brightfield microscopy image}
        \label{subfig:debris_image}
    \end{subfigure}\hfill%
    \begin{subfigure}{.49\linewidth}
        \centering
        \includegraphics[width=\linewidth, frame, trim={0 32cm 0 0cm}, clip]{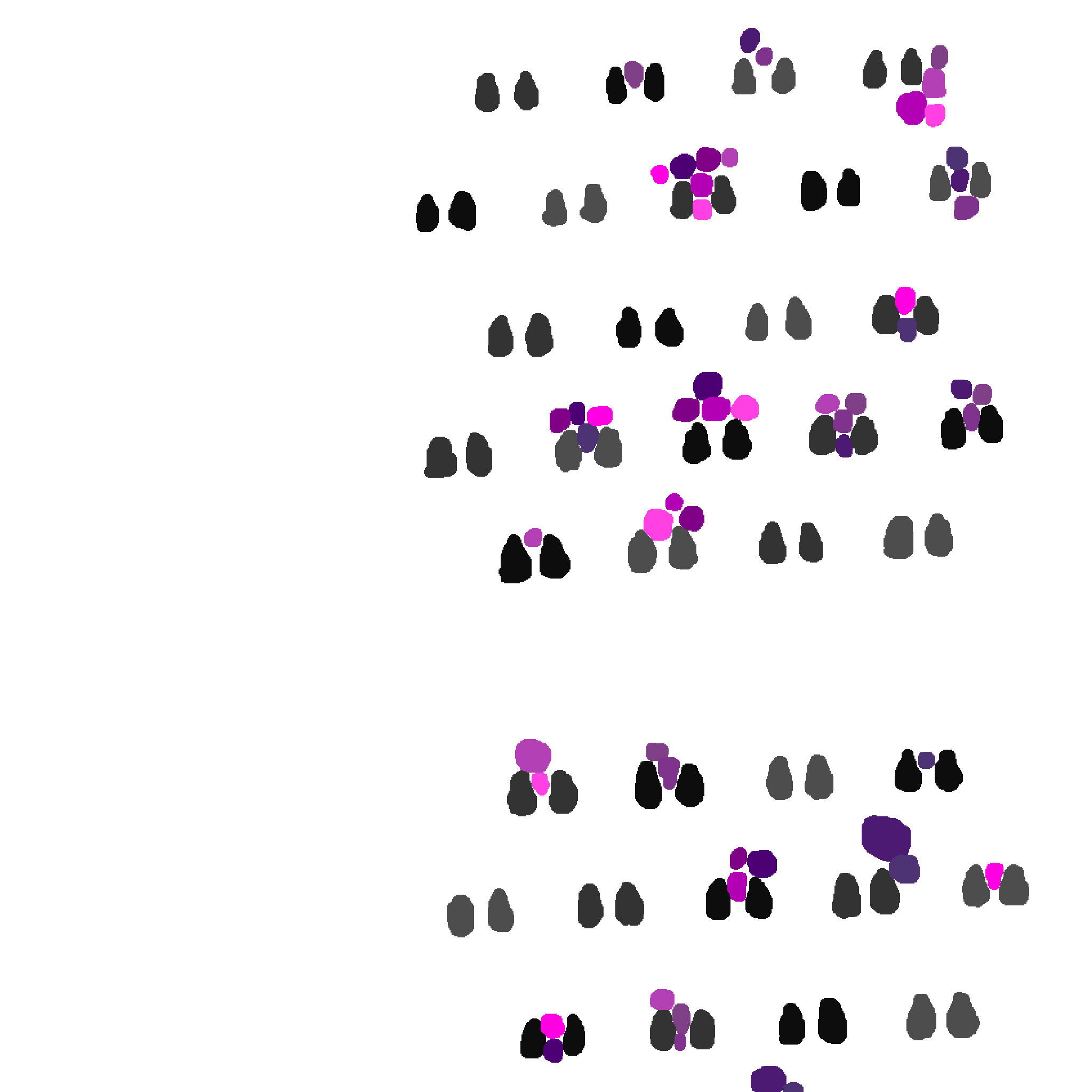}
        \caption{Instance segmentation label}
        \label{subfig:debris_label}
    \end{subfigure}\hfill%
    \caption{\textbf{Strong debris example.} Example of strong debris in the microfluidic chip. For visualization purpose we adjusted the brightness of the microscopy image and show a crop of the full image. Best viewed in color; zoom in for details.}
    \label{fig:debris}
    \vspace{-0.65em}
\end{figure}

We include some challenging edge cases in our labeled dataset, in order to facilitate robust segmentation models. Examples of scenarios where additional objects are in the imagery include debris or contamination on the chip (\cf \cref{fig:debris}). In some TLFM experiments, microchips can get contaminated by debris. In total, our labeled set includes $11$ microscopy images with such challenging edge cases.  Debris adds additional complexity to the microscopy images since debris around cells perceptually affects the appearance of the cell borders.

\subsection{Dataset split} \label{subsec:data_split}

We split our densely instance-wise annotated set into separate sets for training, validation, and testing. For testing, we provide two separate sets. The standard test set partly contains data from experiments that also contributed to the training and validation set. Note while data can be from the same experiment, we do not use the same specific position (\cf \cref{fig:data_acquisition}) for the separate sets. Additionally, we also provide an out-of-distribution (OOD) test set, which only includes images from separate experiments not present in the training, validation, and standard test set. A distribution shift occurs between experiments, due to variations, such as in microchip fabrication or lighting.

Our dataset split is described in \cref{tab:data_split}. While we initially use a random split for the training, validation, and test set, we later curated all three sets to include a representative number of trap types, debris, and focal positions. For the OOD test set, we hand-pick specific samples.

\begin{table}[ht!]
    \centering
    \input{tables/dataset_split}
    \caption{\textbf{Dataset split.} Training, validation, test, and OOD test split of our labeled dataset.}
    \label{tab:data_split}
\end{table}

\subsection{Unlabeled video data} \label{subsec:data_additional}

In addition to the labeled dataset, TYC also includes a large unlabeled dataset, including high-resolution TLFM clips. Motivated by the recent advances in unsupervised optical flow estimation~\cite{Jonschkowski2020, Marsal2023, Meister2018, Ren2017, Stone2021, Wang2018b} and the use of optical flow for tracking~\cite{Decarlo2000, Du2019, Guo2013, Liu2014, Luo2021, Weber2021, Zhang2020}, we aim to facilitate the unsupervised understanding of cell motions with our unlabeled dataset. Our unlabeled dataset might also facilitate general unsupervised representation learning~\cite{Assran2023, Caron2021, Chen2020, Feichtenhofer2022, He2022}. In total, we provide $261$ curated video clips, including $1293$ high-resolution frames.

\begin{figure}[t]
    \centering
    \input{artwork/clip_length_hist}
    \caption{\textbf{Clip lengths histogram.} Frequency of clip lengths.}
    \label{fig:clip_length_hist}
    \vspace{-0.65em}
\end{figure}
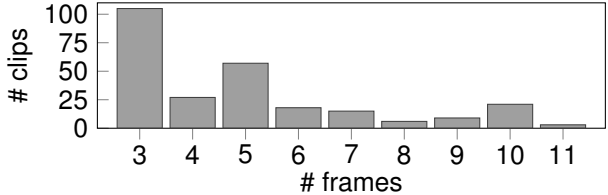

The clip length ranges from $3$ frames to up to $11$ frames with a $\Delta t$ of $10\si{\minute}$. \cref{fig:clip_length_hist} shows a histogram of the different clip lengths. While longer sequences are available, the majority of our clips are composed of three frames.

\begin{figure}[ht!]
    \centering
    \setlength\tabcolsep{1.25pt}
    \begin{tabular*}{\columnwidth}{c c c c c}
        $t_0$ & 
        $t_1$ & 
        $t_2$ & 
        $t_3$ & 
        $t_4$ \\[-0.5pt]
        \includegraphics[width=0.19\linewidth, frame]{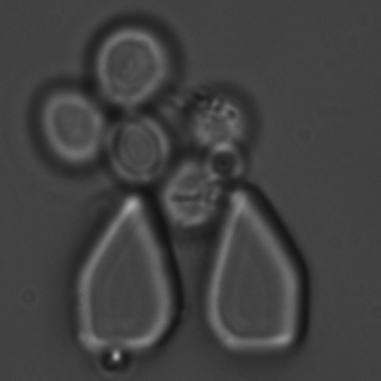} &
        \includegraphics[width=0.19\linewidth, frame]{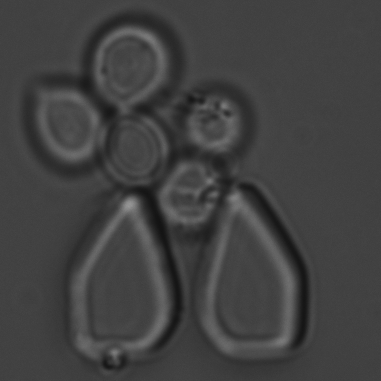} &
        \includegraphics[width=0.19\linewidth, frame]{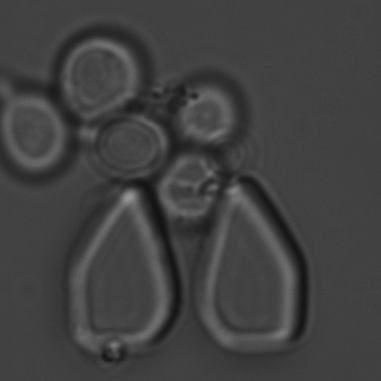} &
        \includegraphics[width=0.19\linewidth, frame]{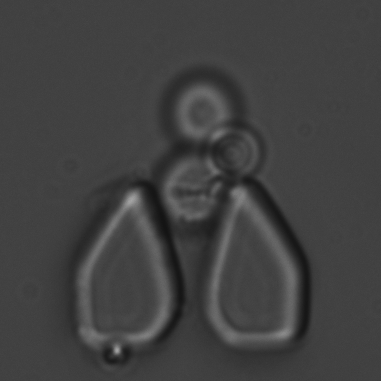} &
        \includegraphics[width=0.19\linewidth, frame]{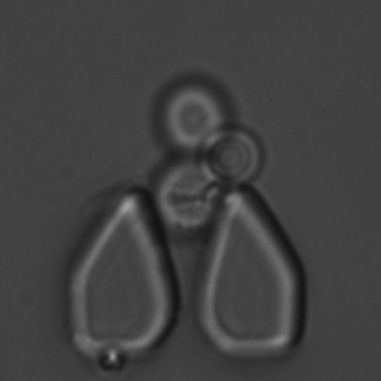} \\[-0.5pt]
        \includegraphics[width=0.19\linewidth, frame]{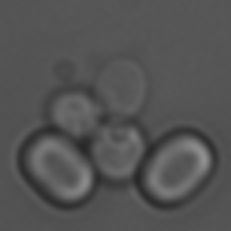} &
        \includegraphics[width=0.19\linewidth, frame]{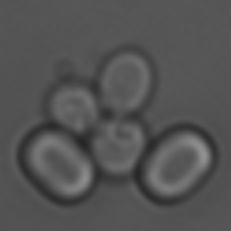} &
        \includegraphics[width=0.19\linewidth, frame]{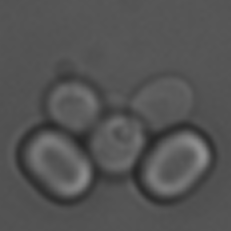} &
        \includegraphics[width=0.19\linewidth, frame]{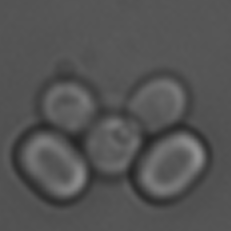} &
        \includegraphics[width=0.19\linewidth, frame]{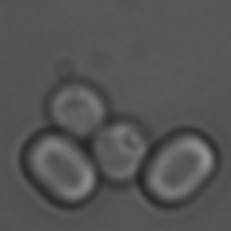} \\
    \end{tabular*}
    \caption{\textbf{Washed out cells.} Example of cells hydrodynamically washed out of a trap pair. In the upper clip, three cells are washed out. In the lower clip, only a single cell is washed away. Frames cropped from the full microscopy clip. $\Delta t$ is $10\si{\minute}$.}
    \label{fig:washed_out_cells}
\end{figure}

We curate our unlabeled dataset; in particular, we ensure no drift of the microfluidic chip. This is to ensure limited global motions, facilitating the learning of fine cell motions. Additionally, we only include clips where no significant amount of cells are washed in or out. While we have no significant amount of washed-out cells, our dataset still includes complex cases of vast cell motions and occasional washed-out cells. Examples of vast cell motions and washed-out cells are shown in \cref{fig:washed_out_cells}.

\begin{figure}[b]
    \centering
    \vspace{-0.65em}
    \setlength\tabcolsep{1.25pt}
    \begin{tabular*}{\columnwidth}{c c c c c}
        $t_0$ & 
        $t_1$ & 
        $t_2$ & 
        $t_3$ & 
        $t_4$ \\[-0.5pt]
        \includegraphics[width=0.19\linewidth, frame]{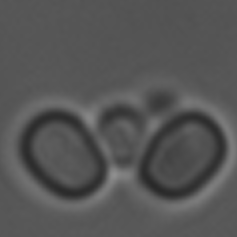} &
        \includegraphics[width=0.19\linewidth, frame]{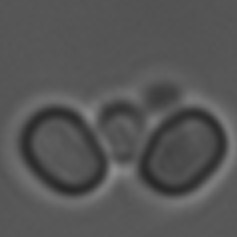} &
        \includegraphics[width=0.19\linewidth, frame]{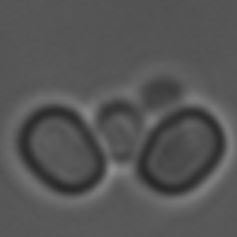} &
        \includegraphics[width=0.19\linewidth, frame]{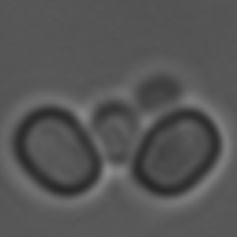} &
        \includegraphics[width=0.19\linewidth, frame]{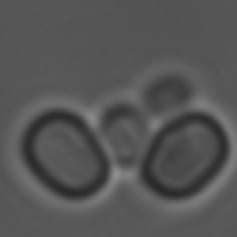} \\[-0.5pt]
    \end{tabular*}
    \caption{\textbf{Budding yeast cell over time.} Temporal sequence of our unlabeled set showcasing budding. A cell grows to the top of the trap. Frames cropped from the full video clip. $\Delta t$ is $10\si{\minute}$.}
    \label{fig:budding_seq}
\end{figure}

Similarly to the labeled dataset, our large unlabeled dataset also includes budding yeast cells. We provide a visualization of the temporal appearance of the budding process in \cref{fig:budding_seq}. The process of budding is, in particular, challenging since small daughter cells appear out of the void due to the large $\Delta t$ of $10\si{\minute}$ between frames.

%% file: artwork/semantic_distribution.tex
\pgfplotsset{
/pgfplots/ybar legend/.style={
/pgfplots/legend image code/.code={\draw[#1, draw] (0cm,-0.1cm) rectangle ++ (0.2cm, 0.2cm);},
}
}
\begin{tikzpicture}[every node/.style={font=\fontsize{10}{10}\sffamily}]
    \begin{groupplot}[
        group style={
            group name=semanticdist,
            group size=2 by 1,
            ylabels at=edge left,
            horizontal sep=5pt,
            vertical sep=5pt,
        },
        ylabel shift=-2.5pt,
        height=3.25cm,
        width=0.59\columnwidth,
        ybar,
        grid=none,
        xtick pos=bottom,
        ytick pos=left,
        ylabel=Proportion,
        xtick={
            0, 1
        },
        xticklabels={
            Cell, Trap
        },
        legend style={nodes={scale=0.5}, at={(0.95, 0.95)}},
        legend image code/.code={\draw[#1, draw] (0cm,-0.1cm) rectangle ++ (0.2cm, 0.2cm);},
        xmin=-0.5, xmax=1.5,
        x tick label style={yshift={0.05cm}},
        ]
        
        \nextgroupplot[title=\vphantom{p}Our dataset, title style={yshift=-8.25pt,}, ybar, bar width=14.2pt, ymax=0.17,ymin=0, ytick={0, 0.05, 0.1, 0.15}, yticklabels={0, 0.05, 0.1, 0.15}, grid=none]
        \addplot[cell, fill, fill opacity=0.5, ybar legend] coordinates { (0.175, 0.0601) };
        \addplot[trap, fill, fill opacity=0.5, ybar legend] coordinates { (0.825, 0.0599) };
        
        \nextgroupplot[title=Reich \etal~\cite{Reich2023}, title style={yshift=-8.25pt,}, ybar, bar width=14.2pt, ymax=0.17, ymin=0, ytick={0, 0.05, 0.1, 0.15}, yticklabels={,,}, grid=none]
        \addplot[cell, fill, fill opacity=0.5, ybar legend] coordinates { (0.175, 0.0827) };
        \addplot[trap, fill, fill opacity=0.5, ybar legend] coordinates { (0.825, 0.1588) };

    \end{groupplot}
\end{tikzpicture}%

%% file: artwork/instance_hist.tex
\begin{tikzpicture}[every node/.style={font=\fontsize{10}{10}\sffamily}]
    \node[anchor=center] at (0.418\columnwidth, -0.675) {\# instances};
    \begin{groupplot}[
        group style={
            group name=foo,
            group size=2 by 1,
            ylabels at=edge left,
            horizontal sep=5pt,
            vertical sep=5pt,
        },
        ylabel shift=-2.5pt,
        height=3.25cm,
        width=0.59\columnwidth,
        ybar,
        major grid style={line width=.2pt,draw=gray!50},
        minor y tick num=1,
        xtick pos=bottom,
        ytick pos=left,
        xmin=0,
        xmax=400,
        ylabel=\# images,
        ymin=0, 
        ymax=15,
        xtick={
            0, 100, 200, 300, 400
        },
        xticklabels={
            , 100, 200, 300, 
        },
        ytick={
            0, 5, 10, 15
        },
        yticklabels={
            0, 5, 10, 15
        },
        scaled x ticks = false,
        legend style={nodes={scale=0.5}, at={(0.975, 0.95)}, align=right},
        legend image code/.code={\draw[#1, draw] (0cm,-0.1cm) rectangle ++ (0.2cm, 0.2cm);},
        x tick label style={yshift={0.05cm}},
        ]
        \nextgroupplot[title=\vphantom{p}Cell, title style={yshift=-8.25pt,}]
        \addplot +[cell, hist={bins=25}, fill opacity=0.5] table [y index=0] {cell_instances.csv};
        
        \nextgroupplot[title=Trap, title style={yshift=-8.25pt,}, xmin=0, xmax=80, yticklabels={,,}, xtick={0, 20, 40, 60, 80}, xticklabels={, 20, 40, 60, }]
        \addplot +[trap, hist={bins=25}, fill opacity=0.5] table [y index=0] {trap_instances.csv};
        
    \end{groupplot}
\end{tikzpicture}%

%% file: artwork/object_size_hist.tex
\begin{tikzpicture}[every node/.style={font=\fontsize{10}{10}\sffamily}]
    \node[anchor=center] at (0.418\columnwidth, -2.575) {\# pixel};
    \node[anchor=center, rotate=90] at (-1.15, -0.1) {\# instances};
    \begin{groupplot}[
        group style={
            group name=foo,
            group size=2 by 2,
            ylabels at=edge left,
            horizontal sep=5pt,
            vertical sep=5pt,
        },
        ylabel shift=-2.5pt,
        height=3.25cm,
        width=0.59\columnwidth,
        ybar,
        major grid style={line width=.2pt,draw=gray!50},
        minor y tick num=1,
        xtick pos=bottom,
        ytick pos=left,
        xmin=0,
        xmax=30000,
        ylabel={},
        xtick={
            0, 5000, 15000, 25000, 30000
        },
        xticklabels={
            , 5k, 15k, 25k, 
        },
        scaled x ticks = false,
        legend style={nodes={scale=0.5}, at={(0.975, 0.95)}, align=right},
        legend image code/.code={\draw[#1, draw] (0cm,-0.1cm) rectangle ++ (0.2cm, 0.2cm);},
        x tick label style={yshift={0.05cm}},
        ]
        \nextgroupplot[title=\vphantom{p}Cell, title style={yshift=-8.25pt,}, ytick={0, 400, 800, 1200}, yticklabels={0, 400, 800, 1200}, xticklabels={,,}, xmin=0, xmax=10000, ymin=0, ymax=1300, xtick={0, 2000, 4000, 6000, 8000, 10000}]
        \addplot +[cell, hist={bins=60}, fill opacity=0.5] table [y index=0] {cell_sizes_new.csv};
        \addlegendentry{Ours}
        
        \nextgroupplot[title=Trap, title style={yshift=-8.25pt,}, ytick={400, 800, 1200}, yticklabels={,,}, ymin=0, ymax=1300, xticklabels={,,}]
        \addplot +[trap, hist={bins=60}, fill opacity=0.5] table [y index=0] {trap_sizes_new.csv};
        \addlegendentry{Ours}
        
        \nextgroupplot[ytick={0, 100, 200, 300, 400}, yticklabels={0, 100, 200, \hphantom{0}300, }, ymin=0, ymax=350, xmin=0, xmax=10000, xtick={0, 2000, 4000, 6000, 8000, 10000}, xticklabels={, 2k, 4k, 6k, 8k, }]
        \addplot +[cell, hist={bins=60}, fill opacity=0.5] table [y index=0] {cell_sizes_old.csv};
        \addlegendentry{Reich~\etal~\cite{Reich2023}}
        
        \nextgroupplot[ytick={0, 100, 200, 300, 400}, yticklabels={,,}, ymin=0, ymax=350]
        \addplot +[trap, hist={bins=60}, fill opacity=0.5] table [y index=0] {trap_sizes_old.csv};
        \addlegendentry{Reich~\etal~\cite{Reich2023}}
        
    \end{groupplot}
\end{tikzpicture}%

%% file: artwork/trap_type_hist.tex
\pgfplotsset{
/pgfplots/ybar legend/.style={
/pgfplots/legend image code/.code={\draw[#1, draw] (0cm,-0.1cm) rectangle ++ (0.2cm, 0.2cm);},
}
}
\begin{tikzpicture}[every node/.style={font=\fontsize{10}{10}\sffamily}]
    \begin{axis}[ 
        ylabel shift=-2.5pt,
        height=3.25cm,
        width=1.0\columnwidth,
        ybar,
        bar width=0.85cm,
        grid=none,
        xtick pos=bottom,
        ytick pos=left,
        ylabel=\# images,
        x tick label style={yshift={0.05cm}},
        xlabel=Trap type,
        xlabel shift=-4pt,
        ymin=0,
        ymax=56,
        ytick={
            0, 15, 30, 45
        },
        yticklabels={
            0, 15, 30, 45
        },
        xtick={
            0, 1, 2, 3, 4, 5, 6
        },
        xticklabel style={text width=0.85cm, align=center},
        xticklabels={
            {1\\[1pt]\includegraphics[width=0.85cm]{artwork/trap_types/0_2.png}}, 
            {2\\[1pt]\includegraphics[width=0.85cm]{artwork/trap_types/1_2.png}}, 
            {3\\[1pt]\includegraphics[width=0.85cm]{artwork/trap_types/2_2.png}}, 
            {4\\[1pt]\includegraphics[width=0.85cm]{artwork/trap_types/3_2.png}}, 
            {5\\[1pt]\includegraphics[width=0.85cm]{artwork/trap_types/4_2.png}},
            {6\\[1pt]\includegraphics[width=0.85cm]{artwork/trap_types/5_2.png}}
        }
        ]
        \addplot[trap, fill, fill opacity=0.5, ybar legend] coordinates { (0, 10) (1, 54) (2, 5) (3, 14) (4, 4) (5, 18) };
    \end{axis}
    
\end{tikzpicture}%

%% file: artwork/cell_density.tex
\begin{tikzpicture}

\definecolor{darkgray176}{RGB}{176,176,176}

\begin{axis}[
colorbar style={
ylabel={},
ytick={1, 21},
yticklabels={L, H}},
colormap={cellmap}{[1pt]
  rgb(0pt)=(1,1,1);
  rgb(1pt)=(0.3,0.2,0.45);
  rgb(2pt)=(1,0.5,0.90980392)
},
xtick=\empty, 
ytick=\empty,
height=4cm,
width=4cm,
point meta max=22,
point meta min=0,
tick align=outside,
tick pos=left,
x grid style={darkgray176},
xmin=0, xmax=2048,
xtick style={color=black},
y dir=reverse,
y grid style={darkgray176},
ymin=-0, ymax=2048,
ytick style={color=black}
]
\addplot graphics [includegraphics cmd=\pgfimage,xmin=0, xmax=2048, ymin=2048, ymax=0] {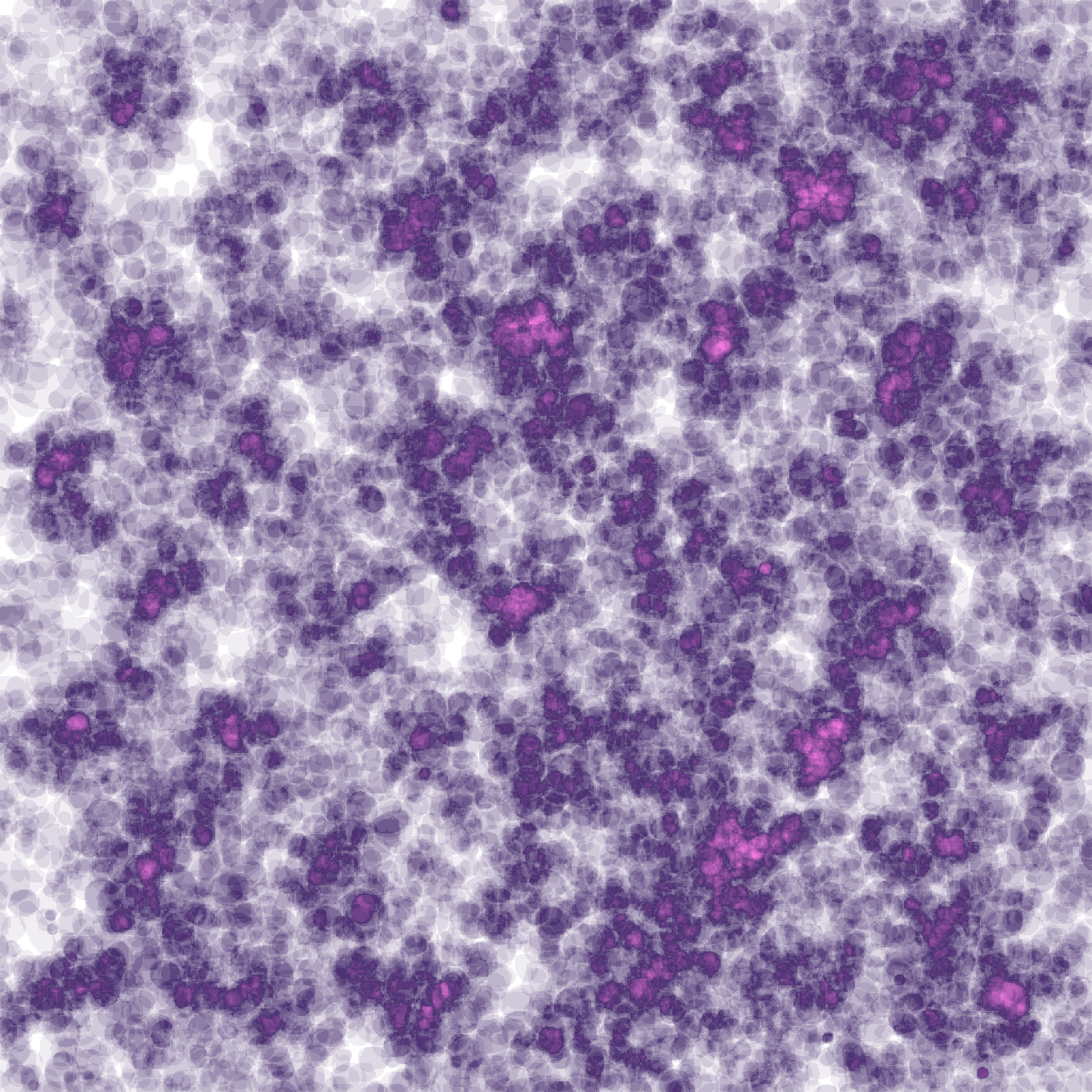};
\end{axis}

\end{tikzpicture}

%% file: artwork/cell_density_old.tex
\begin{tikzpicture}[every node/.style={font=\fontsize{10}{10}\sffamily}]

\definecolor{darkgray176}{RGB}{176,176,176}

\begin{axis}[
colorbar,
colorbar style={
ylabel={},
ytick={20, 329},
yticklabels={L, H}},
colormap={mymap}{[1pt]
  rgb(0pt)=(1,1,1);
  rgb(1pt)=(0.3,0.2,0.45);
  rgb(2pt)=(1,0.5,0.90980392)
},
ticks=none,
xtick=\empty, 
ytick=\empty,
height=4cm,
width=4cm,
point meta max=349,
point meta min=0,
tick align=outside,
tick pos=left,
x grid style={black},
xmin=-0.5, xmax=255.5,
xtick style={color=black},
y dir=reverse,
y grid style={black},
ymin=-0.5, ymax=255.5,
ytick style={color=black},
]
\addplot graphics [includegraphics cmd=\pgfimage,xmin=0, xmax=255.5, ymin=255.5, ymax=-0.5] {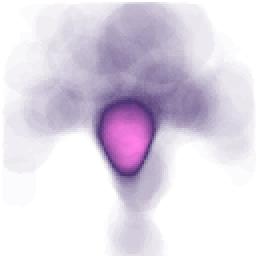};
\end{axis}
\end{tikzpicture}%

%% file: artwork/trap_density.tex
\begin{tikzpicture}

\definecolor{darkgray176}{RGB}{176,176,176}

\begin{axis}[
colorbar style={
ylabel={},
ytick={1, 21},
yticklabels={L, H}},
colormap={trapmap}{[1pt]
  rgb(0pt)=(1,1,1);
  rgb(1pt)=(0.25,0.25,0.25);
  rgb(2pt)=(0.05,0.05,0.05)
},
xtick=\empty, 
ytick=\empty,
height=4cm,
width=4cm,
point meta max=22,
point meta min=0,
tick align=outside,
tick pos=left,
x grid style={darkgray176},
xmin=0, xmax=2048,
xtick style={color=black},
y dir=reverse,
y grid style={darkgray176},
ymin=-0, ymax=2048,
ytick style={color=black}
]
\addplot graphics [includegraphics cmd=\pgfimage,xmin=0, xmax=2048, ymin=2048, ymax=-0.5] {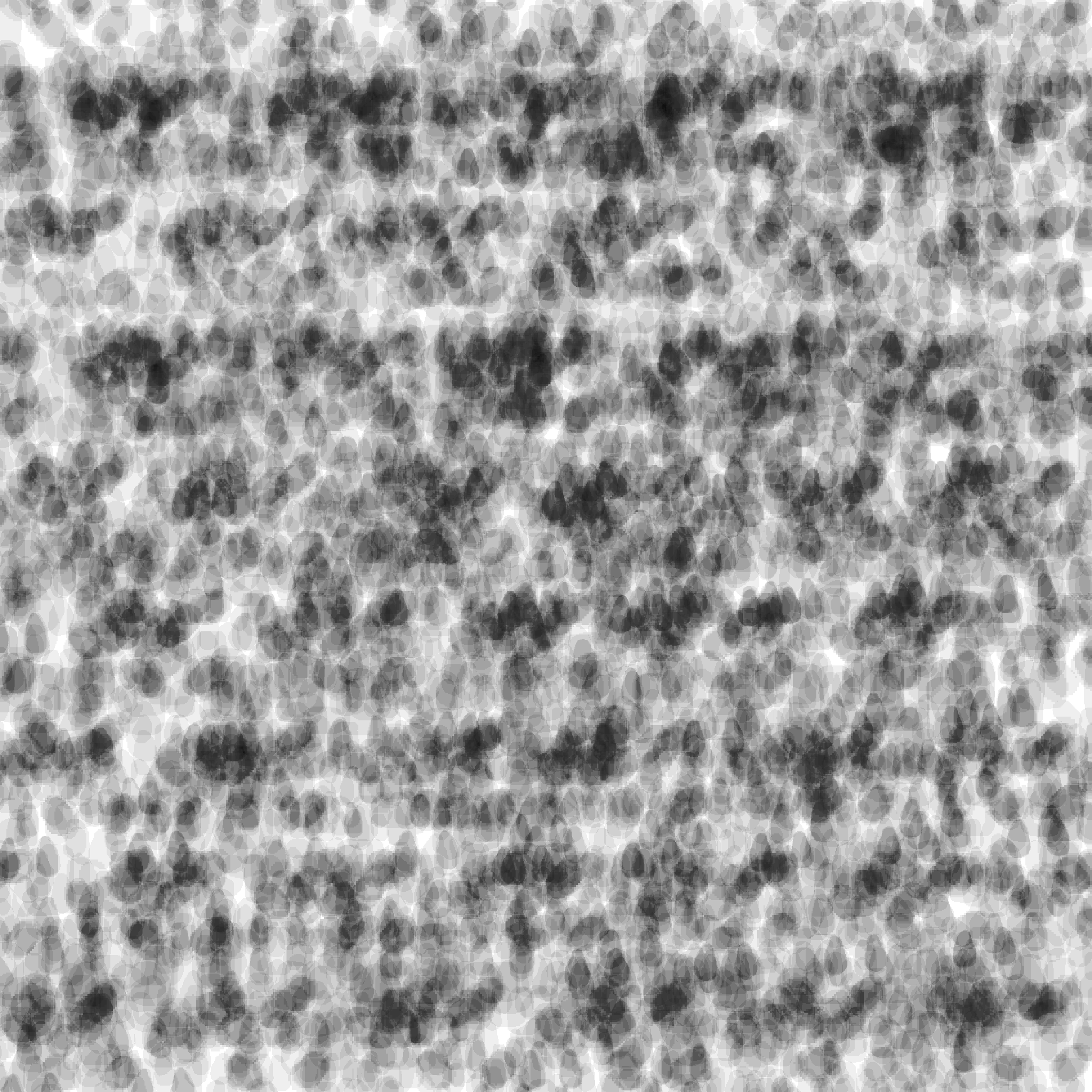};
\end{axis}

\end{tikzpicture}

%% file: artwork/trap_density_old.tex
\begin{tikzpicture}

\definecolor{darkgray176}{RGB}{176,176,176}

\begin{axis}[
colorbar,
colorbar style={
ylabel={},
ytick={1.3, 20.7},
yticklabels={L, H}},
colormap={trapmap}{[1pt]
  rgb(0pt)=(1,1,1);
  rgb(1pt)=(0.25,0.25,0.25);
  rgb(2pt)=(0.05,0.05,0.05)
},
xtick=\empty, 
ytick=\empty,
height=4cm,
width=4cm,
point meta max=22,
point meta min=0,
tick align=outside,
tick pos=left,
x grid style={darkgray176},
xmin=0, xmax=2048,
xtick style={color=black},
y dir=reverse,
y grid style={darkgray176},
ymin=-0, ymax=2048,
ytick style={color=black}
]
\addplot graphics [includegraphics cmd=\pgfimage,xmin=0, xmax=2048, ymin=2048, ymax=-0.5] {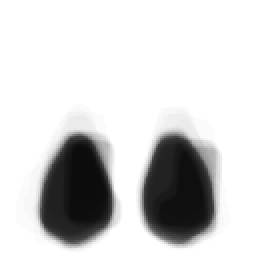};
\end{axis}

\end{tikzpicture}

%% file: tables/dataset_split.tex
{
\small
\begin{tabular*}{\columnwidth}{@{\extracolsep{\fill}}l@{}S[table-format=2.0]S[table-format=2.2]S[table-format=5.0]S[table-format=4.0]}
	\toprule
	Split & {\# images} & {\# ann. pixels [10\textsuperscript{7}]} & {\# cells} & {\# traps} \\
	\midrule
	Training & 81 & 35.53 & 12296 & 3448 \\
	Validation & 8 & 3.69 & 950 & 310 \\
	Test & 8 & 3.59 & 753 & 346 \\
	OOD Test & 8 & 3.58 & 542 & 301 \\
	\bottomrule
\end{tabular*}}

%% file: artwork/clip_length_hist.tex
\pgfplotsset{
/pgfplots/ybar legend/.style={
/pgfplots/legend image code/.code={\draw[#1, draw] (0cm,-0.1cm) rectangle ++ (0.2cm, 0.2cm);},
}
}
\begin{tikzpicture}[every node/.style={font=\fontsize{10}{10}\sffamily}]
    \begin{axis}[ 
        ylabel shift=-2.5pt,
        height=3.25cm,
        width=1.0\columnwidth,
        ybar,
        bar width=0.6cm,
        grid=none,
        xtick pos=bottom,
        ytick pos=left,
        ylabel=\# clips,
        xlabel=\# frames,
        xlabel shift=-4pt,
        ymin=0,
        x tick label style={yshift={0.05cm}},
        ymax=110,
        ytick={
            0, 25, 50, 75, 100
        },
        yticklabels={
            0, 25, 50, 75, 100
        },
        xtick={
            3, 4, 5, 6, 7, 8, 9, 10, 11
        },
        xticklabels={
            3, 4, 5, 6, 7, 8, 9, 10, 11
        }
        ]
        \addplot[trap, fill, fill opacity=0.5, ybar legend] coordinates { (3, 105) (4, 27) (5, 57) (6, 18) (7, 15) (8, 6) (9, 9) (10, 21) (11, 3)};
    \end{axis}
    
\end{tikzpicture}%

%% file: content/evaluation.tex
\section{Evaluation} \label{sec:evaluation}

We propose a standardized evaluation approach for measuring the segmentation performance on our TYC dataset (labeled set). While our dataset offers instance-wise annotations and most biological applications require single-cell information, some biological applications might only rely on semantic-level information (\eg, the fluorescence of all cells). To this end, we offer both semantic and instance-level evaluation approaches. Note our evaluation strategy is strongly inspired by the recent work from Reich~\etal~\cite{Reich2023}.

\subsection{Semantic-level evaluation} \label{sec:evaluation_semantic}

For evaluating the semantic-level performance of segmentation approaches we utilize the cell class intersection-over-union ($\operatorname{IoU}$). The cell class $\operatorname{IoU}$ is computed by:
\begin{equation}
    \operatorname{IoU}\left(p_{\rm c}, g_{\rm c}\right) = \frac{\left|p_{\rm c}\cap g_{\rm c}\right|}{\left|p_{\rm c}\cup g_{\rm c}\right|},
\end{equation}
where $g_{\rm c}$ is the ground truth set of pixels for the cell class and $p_{\rm c}$ denotes the predicted set of cell class pixels. The motivation for utilizing the cell class $\operatorname{IoU}$ is twofold. First, the cell class $\operatorname{IoU}$ is, in particular, significant for biomedical applications only requiring semantic predictions, \eg, measuring the fluorescence of all cells. Second, as previous work (\eg, \cite{Bakker2018} \& \cite{Prangemeier2020b}) reports this metric when evaluating on the dataset by Reich~\etal~\cite{Reich2023} using the cell class $\operatorname{IoU}$ provides a point of comparison to existing work.

\subsection{Instance-level evaluation} \label{sec:evaluation_instance}

We propose to utilize the panoptic quality ($\operatorname{PQ}$) for measuring the instance-level performance on our TYC dataset. In particular, we make use of the property that our instance segmentation labels can also be interpreted as panoptic annotations since we only have one background class and assume no instance overlap (\cf \cref{subsec:data_classes}).

For calculating the $\operatorname{PQ}$, we first match all predicted object masks of a semantic class with the ground truth masks. From this matching, three categories (per semantic class) emerge: True Positive ($\rm TP$), False Positive ($\rm FP$), and False Negative ($\rm FN$) matches. We refer the reader to the work of Kirillov~\etal~\cite{Kirillov2019} for details on the matching approach. After matching, the $\operatorname{PQ}$ is calculated per semantic class (cell, trap, \& background) by:
\begin{equation}
    \small
    \operatorname{PQ}=\underbrace{\frac{\sum_{(p,g)\in {\rm TP}}\operatorname{IoU}(p, g)}{\vphantom{\frac{1}{2}}\left|{\rm TP}\right|}}_{\text{Segmentation quality}\,\operatorname{SQ}}\, \underbrace{\frac{\left|{\rm TP}\right|}{\left|{\rm TP}\right| + \frac{1}{2}\left|{\rm FP}\right| + \frac{1}{2}\left|{\rm FN}\right|}}_{\text{Recognition quality}\,(\operatorname{RQ})},
\end{equation}
where $\frac{1}{\left|{\rm TP}\right|}\sum_{(p,g)\in {\rm TP}}\operatorname{IoU}(p, g)$ calculates the mean intersection-over-union of the matched mask predictions $p$ and the label $g$. After computing the $\operatorname{PQ}$ for each semantic class, the full $\operatorname{PQ}$ is obtained by averaging over all classes.

The $\operatorname{PQ}$ is composed of the segmentation quality ($\operatorname{SQ}$) and recognition quality $\operatorname{RQ}$. This property allows us to separately analyze the segmentation performance and the performance of recognizing objects present in an image. We provide PyTorch~\cite{Paszke2019} code for computing both metrics.

%% file: content/results.tex
\section{SAM Results} \label{sec:sam_results}

SAM (Segment Anything Model) is a recent foundation model for segmentation~\cite{Bommasani2021, Kirillov2023}. Trained on $11\si{\mega\relax}$ diverse images, SAM achieves remarkable zero-shot performance on new image distributions. The support for prompting~\cite{Liu2023} enables SAM to perform diverse segmentation tasks. SAM can be prompted with text, points, boxes, and dense masks. To overcome the ambiguity of segmenting an image, SAM generates multiple mask predictions alongside a score for each prediction. Note SAM only predicts instance masks and not semantic classes, limiting the application of our evaluation strategy. Thus, we provide qualitative results.

Follow-up work has evaluated SAMs zero-shot segmentation performance on biomedical imagery~\cite{Cheng2023, Huang2023, Ma2023, Mazurowski2023}. While SAM achieves strong results on some biomedical tasks (\eg, large-organ seg.) and even exceeds the state-of-the-art, SAM falls short to produce accurate results on other tasks (\eg, pleural effusion seg.)~\cite{Cheng2023, Ma2023}. We utilize SAM to perform zero-shot segmentation of cells and traps.

We employ SAM with a ViT-H backbone~\cite{Dosovitskiy2021, Li2022} to segment cells and traps. We prompt SAM with $256$ points per side (default is $32$) to segment the whole high-resolution microscopy image. To ensure no significant overlap between segmentation masks of individual objects we use a non-maximum suppression IoU threshold of $0.1$ (default is $0.7$). For all other parameters, we utilize the default.

\begin{figure}[b]
    \centering
    \vspace{-0.65em}
    \includegraphics[width=0.31\linewidth, frame]{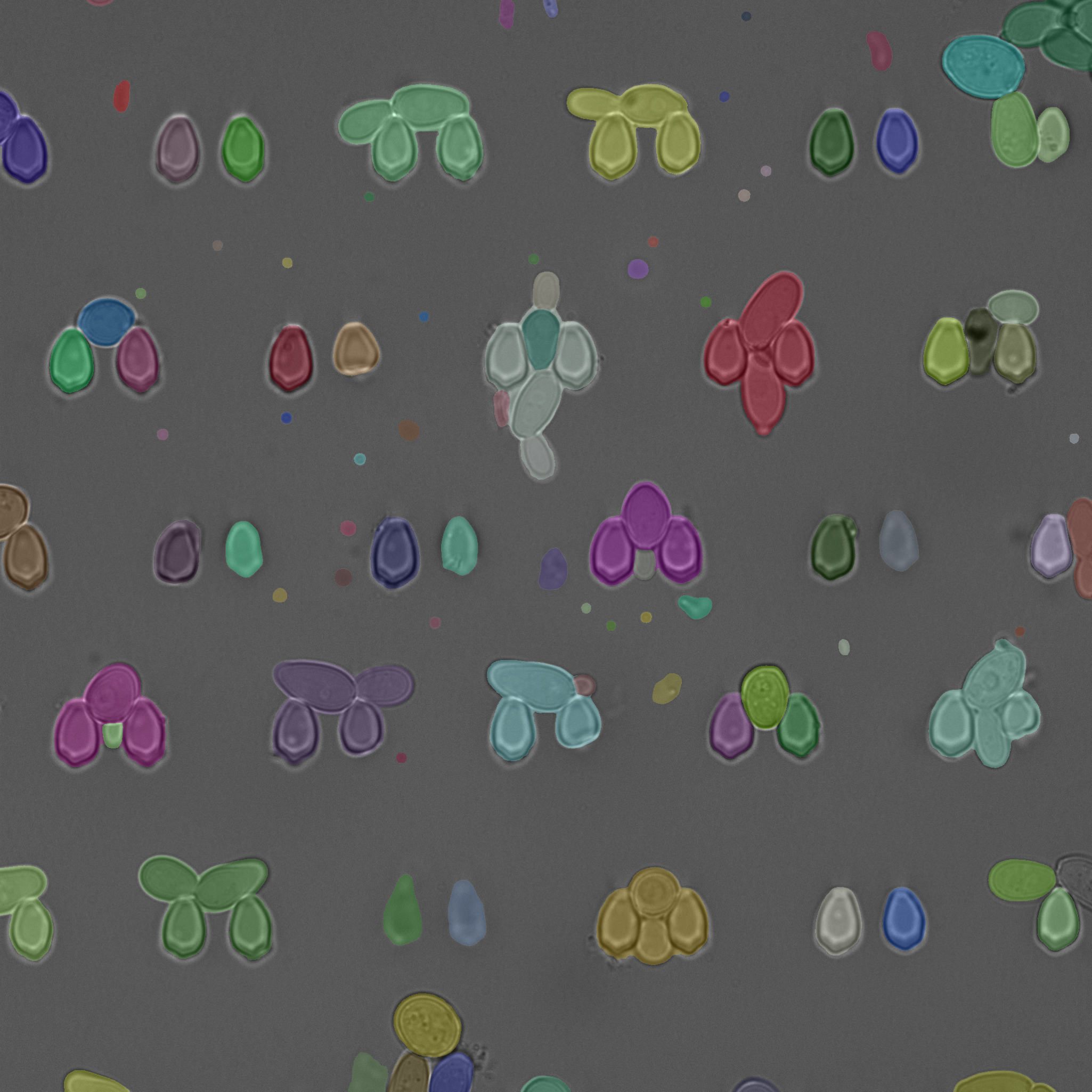}\hfill%
    \includegraphics[width=0.31\linewidth, frame]{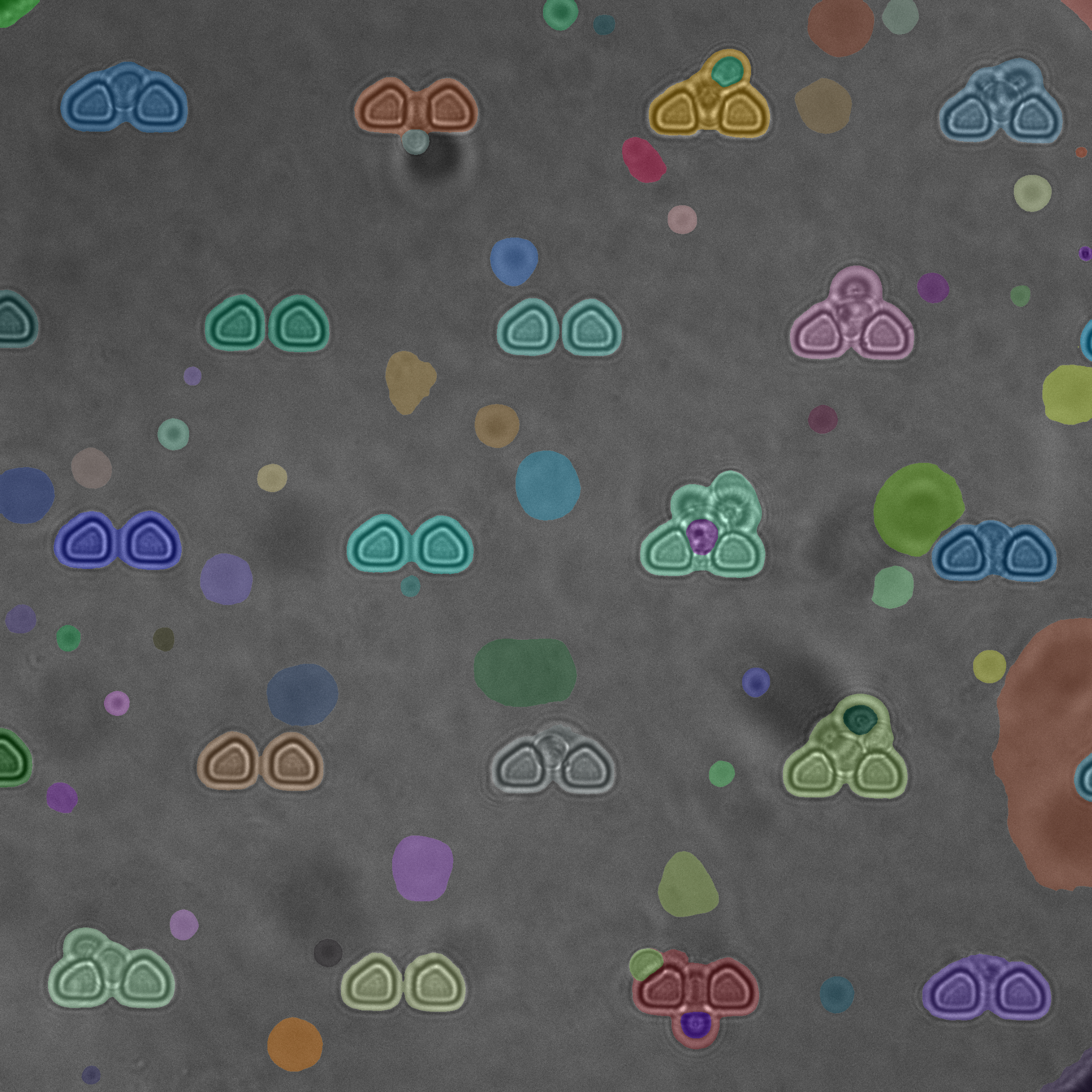}\hfill%
    \includegraphics[width=0.31\linewidth, frame]{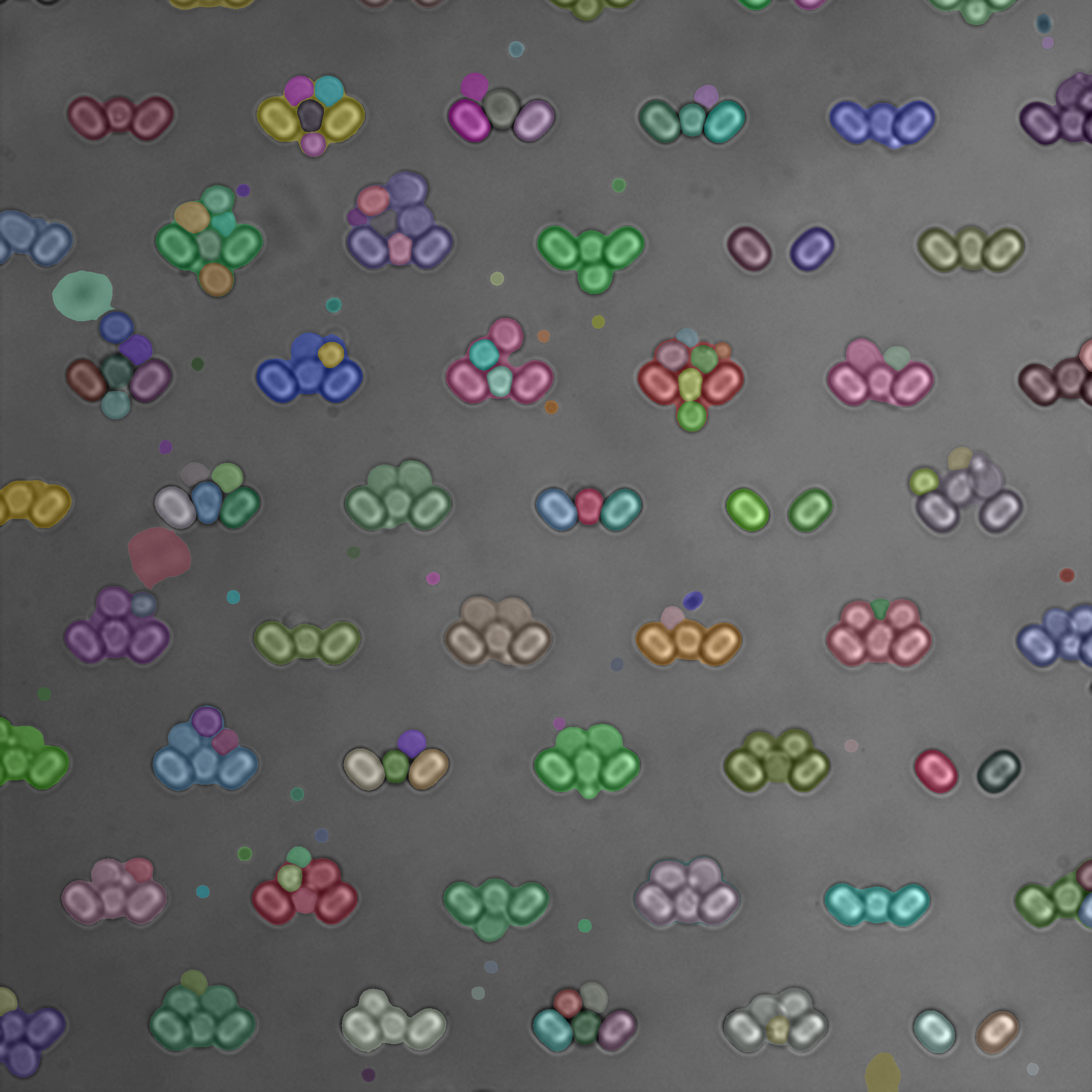}\hfill%
    \caption{\textbf{SAM zero-shot results.} Qualitative segmentation results of SAM on our dataset. Colored masks indicate different instance predictions. Best viewed in color; zoom in for details.}
    \label{fig:sam_results}
\end{figure}

\cref{fig:sam_results} shows quantitative results of SAM on our TYC dataset. In simple cases, SAM yields fairly accurate predictions. In more complex cases (\eg, debris) SAM fails to produce accurate segmentations. We especially observe that touching cells are often segmented as a single object. Additionally, SAM tends to segment background artifacts.

\begin{figure}[t]
    \centering
    \begin{subfigure}{.49\linewidth}
        \centering
        \includegraphics[width=0.49\linewidth, frame]{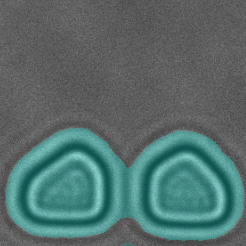}\hfill%
        \includegraphics[width=0.49\linewidth, frame]{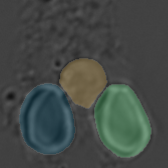}\\[-0.25pt]
        \includegraphics[width=0.49\linewidth, frame]{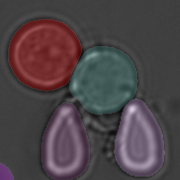}\hfill%
        \includegraphics[width=0.49\linewidth, frame]{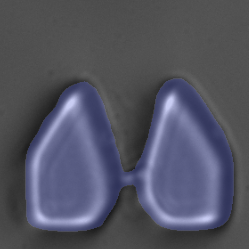}
        \caption{Good examples}
        \label{subfig:sam_good}
    \end{subfigure}\hfill%
    \begin{subfigure}{.49\linewidth}
        \centering
        \includegraphics[width=0.49\linewidth, frame]{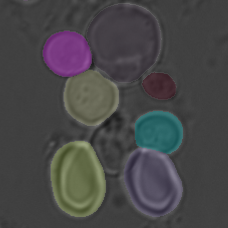}\hfill%
        \includegraphics[width=0.49\linewidth, frame]{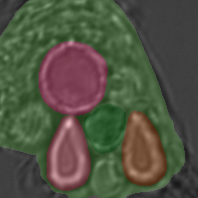}\\[-0.25pt]
        \includegraphics[width=0.49\linewidth, frame]{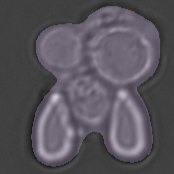}\hfill%
        \includegraphics[width=0.49\linewidth, frame]{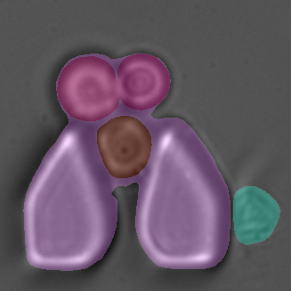}
        \caption{Poor examples}
        \label{subfig:sam_poor}
    \end{subfigure}\hfill%
    \caption{\textbf{Good \& poor SAM results.} \emph{\subref{subfig:sam_good}} In some cases, SAM generates good mask predictions. \emph{\subref{subfig:sam_poor}} In other cases, SAM fails to predict accurate segmentation masks. Color coding as in \cref{fig:sam_results}.}
    \label{fig:sam_results_detail}
\end{figure}

\cref{fig:sam_results_detail} presents both correct mask predictions and failure cases of SAM in more detail. If only a small number of cells are trapped in a single trap pair, SAM typically segments traps and cells correctly (\cf \cref{subfig:sam_good}). Note while we define a pair of traps as a single instance, segmenting both traps separately is technically not wrong due to the ambiguity of zero-shot segmentation. If many cells are trapped in a trap pair, SAM often struggles to detect each cell instance (\cf \cref{subfig:sam_poor}). If an image includes a lot of debris, SAM also struggles to predict correct segmentations.

%% file: content/conclusion.tex
\section{Conclusion and Outlook} \label{sec:outlook}

In this paper, we proposed a novel high-resolution dataset designed to facilitate the understanding of instance-level semantics and motions of yeast cells in microstructures. Our TYC dataset provides pixel-wise instance masks for segmenting yeast cells in microstructures — a widespread task in biological research. Alongside the labeled dataset, we also provide a large unlabeled dataset, including short microscopy video clips to facilitate the unsupervised understanding of cell motions and morphology. To ensure fair comparisons of future cell segmentation approaches, we propose a standardized strategy for evaluating segmentation performance on our dataset. Qualitative zero-shot results of the recent segmentation foundation model SAM demonstrate the complexity of our TYC dataset, as SAM often struggles to generate satisfactory mask predictions. Our effort aims to drive progress in the field of biomedical image analysis and facilitate the development of novel cell segmentation and tracking approaches, as well as make these comparable between laboratories.

While our dataset includes both image and video data, currently only image-level annotations are provided. However, some applications may require temporal labels. We are keen to extend our dataset in the future with video instance segmentation annotations~\cite{Yang2019}, facilitating the development of unified segmentation and tracking approaches.